\documentclass{article}

\usepackage[preprint]{corl_2026}
\makeatother
\usepackage{times}
\usepackage[numbers]{natbib}
\PassOptionsToPackage{bookmarks=true}{hyperref}
\usepackage[dvipsnames, table]{xcolor}
\usepackage{algorithm}
\usepackage{algpseudocode}
\usepackage{amsmath}
\usepackage{bm}

\algrenewcommand\algorithmicrequire{\textbf{Input:}}
\algrenewcommand\algorithmicensure{\textbf{Output:}}
\usepackage{tabularx, booktabs}
\usepackage{multirow}
\usepackage{multicol}
\usepackage{tablefootnote} \usepackage{subcaption}
\usepackage{graphicx}
\usepackage{float}
\usepackage{wrapfig}
\usepackage{amssymb}
\DeclareMathOperator*{\argmax}{argmax}
\DeclareMathOperator*{\argmin}{argmin}
\usepackage[font=small]{caption}
\usepackage{adjustbox}

\usepackage{enumitem}
\usepackage{cleveref}
\usepackage{sidecap}
\usepackage{hyphenat}
\usepackage{makecell}
\usepackage{xspace}

\definecolor{cyangreen}{RGB}{62,154,137}
\newcommand{\shortname}{NovaPlan\xspace}
\newcommand{\method}{NovaPlan\xspace}

\usepackage{longtable}

\renewcommand{\arraystretch}{1.25}
\hypersetup{
  pdftitle={NovaPlan: Zero-Shot Long-Horizon Manipulation via Closed-Loop Video Language Planning},
  pdfauthor={Jiahui Fu, Junyu Nan, Lingfeng Sun, Hongyu Li, Jianing Qian, Benjamin Yang, Yilun Du, Jennifer L. Barry, Kris Kitani, George Konidaris},
  pdfsubject={Robotics, manipulation, and video-language planning},
  pdfkeywords={robot manipulation, video generation, long-horizon planning, zero-shot learning}
}
\usepackage[most]{tcolorbox}

\newtcolorbox{promptbox}[1][]{
  colback=gray!5,
  colframe=gray!40,
  fonttitle=\bfseries,
  coltitle=black,
  title=#1,
  arc=0.5mm,
  boxrule=0.5pt,
  left=1mm, right=1mm, top=1mm, bottom=1mm,
  breakable,
  fontupper=\scriptsize\ttfamily,
}
\title{NovaPlan: Zero-Shot Long-Horizon Manipulation via Closed-Loop Video Language Planning}

\author{
{\setlength{\tabcolsep}{3.5pt}
\begin{tabular}[t]{@{}ccccc@{}}
\textbf{Jiahui Fu$^1$} & \textbf{Junyu Nan$^{1,2}$} & \textbf{Lingfeng Sun$^1$} & \textbf{Hongyu Li$^{1,3}$} & \textbf{Jianing Qian$^4$} \\
\textbf{Benjamin Yang$^5$} & \textbf{Yilun Du$^6$} & \textbf{Jennifer L. Barry$^1$} & \textbf{Kris Kitani$^2$} & \textbf{George Konidaris$^3$}
\end{tabular}} \\
\mdseries\small $^1$Robotics and AI Institute \quad $^2$Carnegie Mellon University \quad $^3$Brown University \\
\mdseries\small $^4$University of Pennsylvania \quad $^5$University of Illinois Urbana-Champaign \quad $^6$Harvard University}

\begin{document}
\raggedbottom
\maketitle
\vspace{-1em} 
\begin{abstract}
Solving complex long-horizon robotic tasks requires joint reasoning over abstract task structure and low-level physical interaction. While combining Vision-Language Models (VLMs) and video generation models offers a promising path for zero-shot planning, their individual tendencies to hallucinate physics or violate geometric consistency often compound over time, preventing reliable real-world execution. We introduce NovaPlan, a hierarchical framework that enables robust, zero-shot long-horizon manipulation by systematically proposing, verifying, and repairing visual plans. At the high level, a VLM planner decomposes tasks and filters out dynamically inconsistent futures by verifying multiple candidate video rollouts. To translate these imagined futures into reliable physical actions, NovaPlan utilizes a hybrid geometric representation that adaptively switches between object-centric flow and human hand flow. Finally, NovaPlan closes the loop by continuously monitoring execution to verify outcomes and synthesize local, non-prehensile corrective behaviors, such as fingertip poking, when failures occur. Across diverse multi-stage tasks, NovaPlan substantially outperforms prior zero-shot systems, achieving complex assembly and dexterous error recovery entirely without task-specific training or demonstrations. Please visit our project website for additional results: \href{https://nova-plan.github.io/}{\textcolor{orange}{https://nova-plan.github.io/}}.
\end{abstract}

\keywords{Long-horizon planning, Zero-shot manipulation, Foundation models} 

\section{Introduction}
\label{sec:intro}
A central challenge in robotics is enabling agents to solve complex long-horizon tasks. Solving such tasks requires an agent to jointly reason about
\emph{abstract task structure}, deciding what sequence of actions should achieve a goal as well as \emph{physical interaction}, deciding how objects, contacts, and constraints evolve under execution.  
These two forms of reasoning are tightly coupled: an action that is semantically correct may still make later interactions infeasible if it blocks access, causes misalignment, or leaves the scene in an unexpected state. Classical robotic systems typically separate these problems: symbolic planners reason over high-level task abstractions, while low-level controllers operate on geometric or dynamical models~\citep{cambon2009hybrid, wolfe2010combined, kaelbling2011hierarchical}. However, such modular pipelines struggle to scale to open-world environments due to extensive task- and object-specific modeling.

Recent foundation models suggest an alternative modular decomposition. Vision-language models (VLMs)
have demonstrated strong semantic understanding and high-level reasoning capabilities~\citep{singh2026openaigpt5card, comanici2025gemini}, while
video generation models increasingly exhibit implicit knowledge of object dynamics, contact,
and temporal evolution~\citep{wan_Wan_2025, wiedemer_Video_2025, team_Evaluating_2026}. By jointly planning using both VLMs and video models through visual imagination~\citep{ajay2023compositional, du_Video_2023}, we can combine both high- and low-level knowledge and construct a long-horizon robot planning system that can reason about both \emph{abstract task structure} and \emph{physical interaction}. In principle, this allows a robot to plan over imagined object interactions without task-specific demonstrations or manually specified dynamics.

\begin{figure}[t]
    \includegraphics[width=\linewidth]{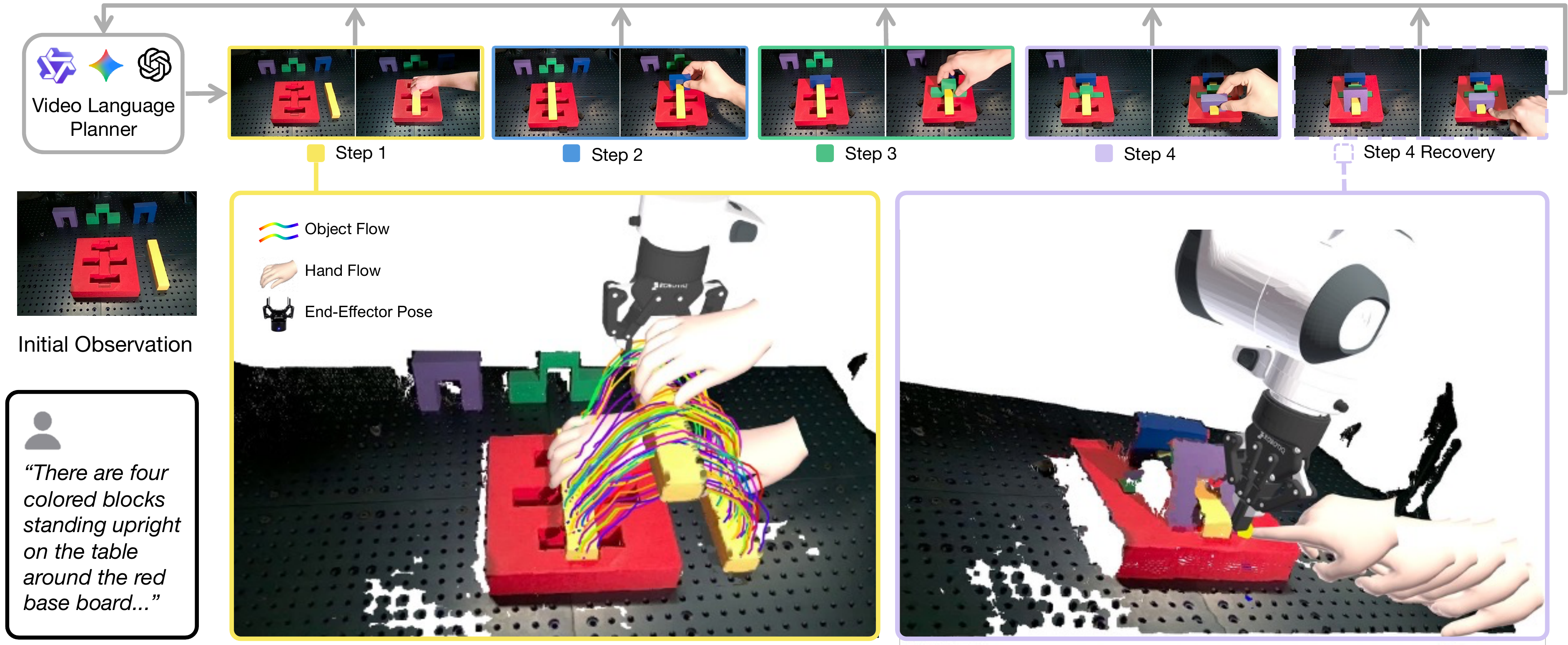}
    \caption{
        \textbf{Zero-shot long-horizon manipulation with NovaPlan.} Given a task command and an initial observation, a video language planner is used to select and generate the best next-step video. Executable robot actions are extracted from the video through object or hand flow tracking. The updated observation is sent back to the planner to enable closed-loop reasoning and recovery from failure states.
    }
    \label{fig:teaser}
\end{figure}

However, generated visual futures are not reliable transition models. 
VLMs may propose semantically plausible but physically infeasible actions, while video models may hallucinate objects and violate geometry. These errors are especially damaging in long-horizon manipulation, where deviations in imagined states can compound into full-task failure. Robust long-horizon operation therefore requires a closed-loop procedure that can effectively recover from these failures.

We present \textbf{NovaPlan}, a closed-loop framework for robustly solving zero-shot long-horizon manipulation tasks, where video planning is treated as a structured proposal mechanism within an online verification and repair system.  At a high level, NovaPlan combines semantic task reasoning from
VLMs with visual future prediction from video generation models, while augmenting these components with explicit verification, adaptive geometric grounding, and closed-loop recovery, enabling reliable execution of long-horizon tasks in a zero-shot manner.

\looseness=-1
NovaPlan robustly solves long-horizon tasks through three mechanisms. First, to construct
\textit{robust high-level planning}, the system generates multiple candidate video plans, verifies each of them using structured semantic and physical consistency checks, and filters out implausible futures from both VLM and video models. Second, to enable
\textit{geometrically grounded execution} from plans, NovaPlan converts generated videos into robot trajectories
through a hybrid representation that adaptively switches between object-centric flow and hand-centric
flow, enabling us to robustly infer executable actions across videos.
Third, to enable \textit{closed-loop execution and recovery from inaccurate plans}, NovaPlan continuously re-grounds and regenerates plans from current observations, verifies successful execution, and synthesizes local
corrective behaviors, including non-prehensile recovery actions, when failures occur. 

We evaluate NovaPlan on diverse long-horizon manipulation tasks and the challenging Functional Manipulation Benchmark (FMB). 
Across multi-step tasks including block stacking, object sorting, and exploratory hidden-object search, NovaPlan shows substantially improved robustness over various prior zero-shot manipulation approaches~\cite{du_Video_2023,li_NovaFlow_2025, liu_MOKA_2024, ye_DreamZero_2026,intelligence_$p_05$_2025}. 
NovaPlan also demonstrates zero-shot local recovery behaviors, such as fingertip poking for object realignment.

Overall, our contributions are as follows: 
\textbf{(1)} We present NovaPlan, a closed-loop video language planning framework that enables robust long-horizon task execution by systematically proposing, verifying, and re-grounding visual plans. \textbf{(2)} We develop a hybrid visual grounding module that converts accepted visual transitions into executable robot trajectories by adaptively switching between object-centric flow and hand-centric flow. 
\textbf{(3)} We validate NovaPlan on diverse multi-step manipulation tasks, where it outperforms various leading zero-shot baselines.

\section{Related Work}
\label{sec:related}
\textbf{Zero-shot manipulation.}
Inspired by recent advances in Large Language Models (LLMs) and Vision-Language Models (VLMs)~\cite{qwen_Qwen25_2025,touvron_llama_2023,liu_visual_2023}, there is growing interest in solving robotic manipulation tasks through zero-shot generalization.
One line of work trains Vision-Language-Action (VLA) models~\cite{kim_OpenVLA_2025,ghosh_Octo_2024,team_Careful_2025} on large-scale real-world robot trajectories to learn generalizable control policies.
Several recent VLAs further inject 3D structure into the policy backbone to improve geometric precision~\cite{sun_GeoVLA_2025,qu_SpatialVLA_2025,li_3DSVLA_2025,huang_GraphCoTVLA_2025}, although that structure is learned inside a policy trained on robot demonstrations.
VLA scalability also remains limited by the cost of collecting physical interaction data relative to web-scale text and images.
An alternative line of work uses VLMs to generate symbolic or spatial plans rather than direct motor commands~\cite{liu_MOKA_2024,huang_ReKep_2025,du_Video_2023,yin_CodeDiffuser_2025, liu2025simpact}, including agents that emit robot code against a designer-provided API~\cite{fu_CaPX_2026}.
While effective at capturing task structure, these methods often lack the fine-grained 3D spatial awareness required for robust manipulation.

\textbf{Video generation for robotics.}
Video generation models provide a promising representation for dynamic reasoning and future-state prediction, which are leveraged by robotic applications in several distinct ways.
One family of work predicts future frames from an initial observation and a language instruction (image-to-video, I2V), which requires no action labels at inference time.
Other work instead conditions video generation on robot actions, yielding controllable world models~\cite{guo_CtrlWorld_2025}, simulators for policy evaluation~\cite{quevedo_WorldGym_2026}, and world action models that act as zero-shot policies~\cite{ye_DreamZero_2026,yuan_FastWAM_2026}.
While this enables tighter controllability, it requires action-annotated data and ties the model to its training embodiments.
Prior work converts generated videos into robot actions using inverse dynamics models~\cite{du_Learning_2023,ajay2023compositional,luo_Solving_2024} or imitation policies~\cite{bharadhwaj_gen2act_2024,li_Unified_2025,liang_Video_2025}, both of which require task-specific demonstrations.
Other work extracts control signals directly using 6D pose estimation~\cite{patel_Robotic_2025,yin_Objectcentric_2025} or flow tracking~\cite{li_NovaFlow_2025,bharadhwaj_gen2act_2024}.
However, these approaches remain sensitive to perception error, tracking drift, and open-loop execution, which limits robustness over long horizons.
Closest to our setting, video language planning~\cite{du_Video_2023} and compositional hierarchical planning~\cite{ajay2023compositional} search jointly over language actions and imagined video futures, and our beam search follows their search method.
VLP prompts a VLM to measure progress to final goals, and notes in its limitations that this often leads to inconsistent video plans.
\shortname instead applies target, physical, motion, and result checks, converts videos into metric robot motion through object- or hand-flows with no robot training, and re-grounds each plan on the latest visual observation with local recovery rather than receding-horizon replanning, which can leave later actions conditioned on stale predicted states (\Cref{fig:long_horizon_result}).

\vspace{-7pt}
\section{Method}
\vspace{-7pt}
\label{sec:method}
\begin{figure*}[t]
    \centering
    \includegraphics[width=\linewidth]{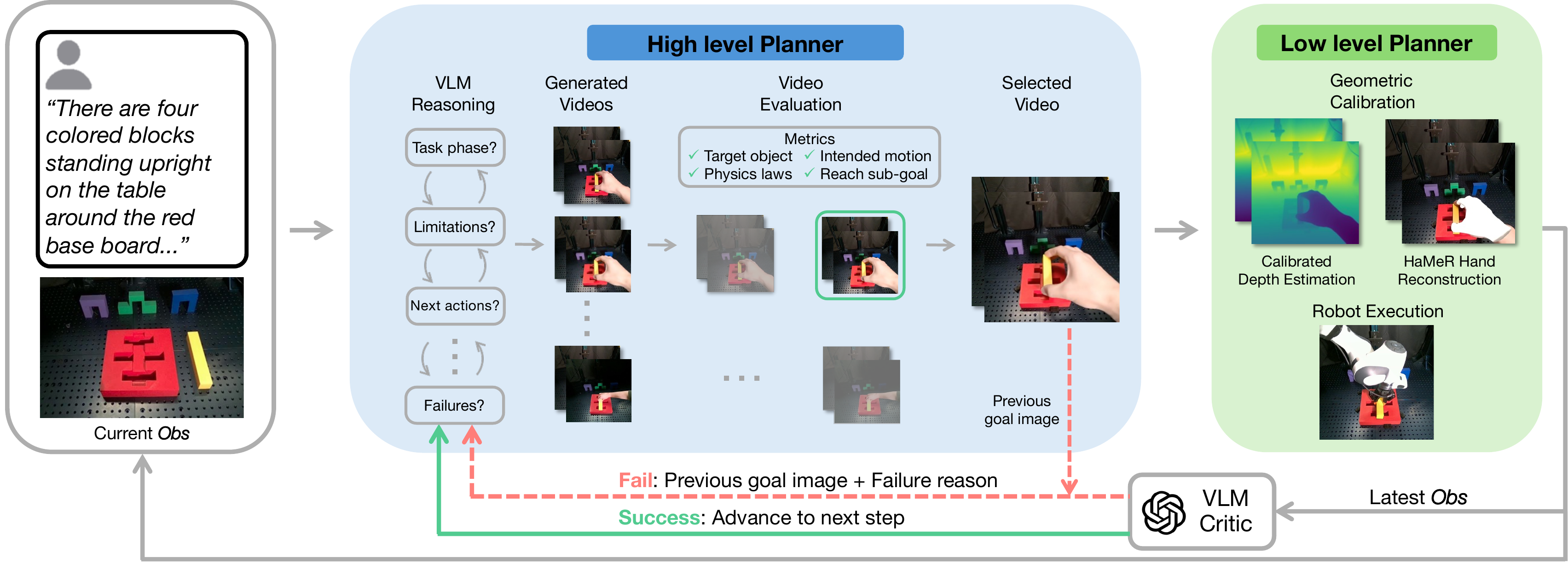}
    \caption{\textbf{\method system overview.}
    A high-level planner takes in the task instruction and current observation, and proposes multiple task-solving videos after VLM reasoning about the task progress. Videos are selected based on flow and semantic consistency. The low-level planner calculates the robot action from the extracted hand or object flow. Updated observations are sent to a VLM critic to determine whether the robot should proceed to the next step or recover from failure.
    }
    \label{fig:system}
\vspace{-10pt}
\end{figure*}
Our goal is to develop a closed-loop hierarchical video language planner for zero-shot long-horizon manipulation.
As shown in \Cref{fig:system}, NovaPlan treats generated videos as visual transition proposals inside an online planning, verification, and repair loop.
The planner proposes candidate macro-actions, samples video rollouts, verifies their semantic and physical consistency, and re-grounds selected transitions on the latest real observation before execution (\Cref{sec:video_planning}).
Accepted visual transitions are then converted into executable robot trajectories through a hybrid object-flow and hand-flow grounding module (\Cref{sec:grounding}).

\vspace{-5pt}
\subsection{Closed-Loop Video Language Planning} 
\vspace{-5pt}
\label{sec:video_planning}
NovaPlan's high-level planner combines VLM-based text action generation with video-based physical verification to reason about possible future actions. The planner uses a proposal--rollout--verification procedure: the VLM proposes one or more candidate text actions from a source image and task context, the video model instantiates each action as multiple visual rollouts, and a VLM critic filters physically or semantically inconsistent futures to choose the most plausible rollout for execution. This procedure is used inside strategic beam search to build full text plans for coupled tasks, reactively to select the next action for exploratory or non-coupled tasks, and locally to produce recovery actions after execution failures. As summarized in Algorithm~\ref{alg:closed_loop_vlp}, NovaPlan closes the loop by verifying real execution outcomes and invoking local recovery when the observed transition deviates from the intended visual target. Please see Appendix~\ref{app:VLM_planning} for full design details.

\begin{wrapfigure}{r}{0.58\linewidth}
\vspace{-20pt}
\begin{minipage}{\linewidth}
\begin{algorithm}[H]
\caption{Closed-loop video language planning}
\label{alg:closed_loop_vlp}
\small
\begin{algorithmic}[1]
\Require goal $g$, initial RGB $I^{\mathrm{real}}_0$, horizon cap $N$, video model $G$
\State $I\gets I^{\mathrm{real}}_0$, $\mathcal{H}\gets\emptyset$, $\xi\gets\emptyset$, $\mathcal{Q}\gets\emptyset$
\State $m\gets\textsc{AssessTaskStructure}(I,g)$ \Comment{strategic or reactive}
\If{$m=\mathrm{strategic}$}
    \State $\mathcal{Q},\xi\gets\textsc{StrategicBeamSearch}(G,I,g,\mathcal{H},\xi,N)$
\EndIf
\While{task not complete}
    \If{$m=\mathrm{strategic}$}
        \If{$\mathcal{Q}=\emptyset$}
            \State \textbf{break} \Comment{planned sequence completed}
        \EndIf
        \State $(a,o)\gets\textsc{Pop}(\mathcal{Q})$
    \Else
        \State $(a,o,\xi)\gets\textsc{ProposeAction}(I,g,\mathcal{H},\xi)$
    \EndIf
    \State $V\gets\textsc{GenerateAndSelectRollout}(G,I,a,o)$
    \State $I^{\mathrm{target}}\gets\mathrm{LastFrame}(V)$
    \State $\tau\gets\textsc{GroundVideo}(V,o)$; execute $\tau$ and observe $I^+$
    \If{$\textsc{VerifyTransition}(I,I^+,I^{\mathrm{target}},a)$}
        \State $\mathcal{H}\gets\mathcal{H}\cup\{a\}$; $I\gets I^+$
    \Else
        \State $r,I^+\gets\textsc{AttemptRecovery}(G,I^+,I^{\mathrm{target}},a,g)$
        \If{$r=\mathrm{success}$}
            \State $\mathcal{H}\gets\mathcal{H}\cup\{a\}$; $I\gets I^+$
        \Else
            \State \textbf{break} \Comment{recovery failed}
        \EndIf
    \EndIf
\EndWhile
\end{algorithmic}
\end{algorithm}
\end{minipage}
\vspace{-18pt}
\end{wrapfigure}

\textbf{High-level action proposal.}
Given a source image $I_{\mathrm{src}}$, task goal $g$, action history $\mathcal{H}$, and constraint context $\xi$, the VLM proposes candidate macro-actions 
$\Pi=\{(a^i,o^i,\xi^i)\}_{i=1}^{N_c}$,
where $a^i$ is a language-level manipulation action, $o^i$ is the object or region to track, and $\xi^i$ updates reasoning over task progress. $I_{\mathrm{src}}$ is the last frame of the previously selected video during strategic planning and the latest real observation during reactive planning, execution, and recovery. NovaPlan proposes multiple actions each step in strategic planning to  reason over future action orders, and only the next action in reactive planning and recovery for efficiency.

\looseness=-1
\textbf{Video rollouts as physical tests.}
The proposed actions are next converted into imagined visual futures. 
For each action $a^i$, NovaPlan samples $L$ rollouts from the video model,
$\{V^{i,k}\}_{k=1}^{L}=G(I_{\mathrm{src}},a^i)$.
Each rollout is treated as a hypothesis about what the action would physically do from the given source image. This same rollout-generation step is used in three places: in strategic beam search to imagine future states, in execution to instantiate the selected text command from the latest real scene, and in recovery to generate candidate corrective motions toward the previous target state.

\textbf{Hierarchical rollout verification.} As generated videos can look plausible while being physically wrong, NovaPlan verifies each rollout before using it for planning, execution, or recovery. As described in Appendix~\ref{sec:video_selection}, a VLM critic applies four ordered checks: \emph{target consistency}, whether the intended object moves; \emph{physical consistency}, whether the video avoids artifacts such as deformation or hallucinated objects; \emph{motion consistency}, whether the tracked flow matches the action command; and \emph{result consistency}, whether the final frame achieves the subgoal. This hierarchy reduces hallucination in the VLM critic and filters out goal-like final frames reached via impossible or irrelevant motions. 
NovaPlan batch-scores rollouts by arranging each candidate's final frame and its flow image (object flow across the rollout) into a grid, enabling efficient comparison of multiple futures. 

\textbf{Verified lookahead.}
The planning horizon is selected based on the structure of the task. For exploratory tasks, such as opening drawers to search for a hidden object, the next useful action depends on information revealed by execution. 
In this case, NovaPlan uses a greedy horizon $h=1$, executes one verified action, and observes again. 
For coupled tasks, such as assembly, the order of actions can determine whether later insertions remain feasible. 
In this case, NovaPlan uses a strategic horizon $h=N$ and performs beam search~\cite{du_Video_2023} in advance to generate the full task plan to be followed during execution, as detailed in Appendix~\ref{sec:tree_search}. 
Each beam stores a starting image, an action history, a constraint context, and a rollout score. The final frame of the last selected rollout becomes the starting image for the next search depth, allowing video generation to serve as a coarse object-dynamics model to reason over action orders for full-step planning before execution.

\textbf{Execution-time re-grounding.}
During execution, NovaPlan follows the text action command proposed by the planner: for strategic tasks, the command is popped from the beam-search plan, while for reactive tasks it is proposed from the current observation. 
Before executing the selected action $a_t$, NovaPlan generates candidate videos from the latest real observation,
$\{V^k_t\}=G(I^{\mathrm{real}}_t,a_t)$.
The same rollout verifier selects the execution video $V^{\mathrm{exec}}_t$, and its last frame defines the local target state,
$I^{\mathrm{target}}_t=\mathrm{LastFrame}(V^{\mathrm{exec}}_t)$.
This re-grounding step ensures that the low-level planner always receives a visual transition aligned with the current scene rather than relying on an imagined starting condition produced earlier during planning.

\textbf{State verification after execution.}
After the grounded trajectory is executed, NovaPlan checks whether the real scene undergoes the intended transition. As detailed in Appendix~\ref{app:state_verification}, the VLM critic compares three images: the pre-execution observation $I^{\mathrm{real}}_t$, the post-execution observation $I^{\mathrm{real}}_{t+1}$, and the generated target frame $I^{\mathrm{target}}_t$, 
and determines if
$I^{\mathrm{real}}_t \rightarrow I^{\mathrm{real}}_{t+1}$
matches the planned visual transition
$I^{\mathrm{real}}_t \rightarrow I^{\mathrm{target}}_t$.
If verification succeeds, the planner advances to the next action. 
If verification fails, the failure explanation is passed to the recovery stage.

\textbf{Local recovery.} When execution fails, NovaPlan attempts a local repair operation. 
The previous target frame $I^{\mathrm{target}}_t$ remains the intended local goal, while the current observation $I^{\mathrm{real}}_{t+1}$ becomes the new start state. 
As described in Appendix~\ref{app: recovery_proposal}, the VLM compares these two images given the intended action, selects a recovery mode
$S_t\in\{\mathrm{grasp},\mathrm{non\mbox{-}prehensile}\}$, and proposes a recovery action following the selected mode. 
The same proposal--rollout--verification pipeline is then used to generate and select a recovery video for local repair. 
If the recovered state reaches the previous target, NovaPlan resumes the high-level plan; otherwise, the recovery attempt is marked as failed.

\subsection{Geometric Grounding of Verified Video Plans}

\label{sec:grounding}
During execution, after the high-level planner selects a verified video transition, NovaPlan grounds the visual motion into an executable end-effector trajectory. 
As the generated videos may contain geometric artifacts and are not metric robot trajectories, NovaPlan extracts motion by leveraging two complementary 3D cues: object flow and hand flow. 
Object flow recovers target object motion when the object remains visible and rigidly trackable, while hand flow uses the generated human hand as a kinematic prior when object tracking fails due to occlusion, large rotations, or unstable depth. 
This hybrid design allows NovaPlan to robustly convert imperfect generated videos into useful robot motion. Please see Appendix~\ref{app:grounding_details} for full design details.

\textbf{Object-centric grounding.}
For object-centric grounding, we recover a metric 3D trajectory of the target object from the generated video, following the general actionable flow pipeline of NovaFlow~\cite{li_NovaFlow_2025}. As described in Appendix~\ref{app:object_flow}, given a generated video and the target object specified by the planner, we first estimate per-frame geometry using a monocular geometry model~\cite{wang_MoGe2_2025}, optionally refine the depth sequence for temporal consistency using CVD~\cite{luo_Consistent_2020}, and align the initial generated depth to the real sensor depth through affine calibration~\cite{patel_Robotic_2025,li_NovaFlow_2025}.
We then segment the target object, sample object points, and track them in 3D using a dense point tracker~\cite{zhang_TAPIP3D_2025}.
The resulting object flow is denoted as
\begin{equation*}
    \mathcal{F}_{\mathrm{obj}} = \{ \mathbf{f}_i^t \in \mathbb{R}^3 
    \mid i=1,\dots,K,\; t=1,\dots,T \},
\end{equation*}
where $\mathbf{f}_i^t$ is the 3D position of object point $i$ at frame $t$.
We fit a rigid transform between consecutive frames from the tracked points, yielding an object motion sequence $\{T_{\mathrm{obj}}^t\}_{t=1}^{T}$.

To execute the object motion, we generate a grasp on the initial object geometry using a grasp proposal model~\cite{murali_GraspGen_2025}.
The selected grasp defines a fixed object-to-end-effector transform $T_{\mathrm{obj}\rightarrow \mathrm{ee}}$.
Assuming the object remains fixed relative to the gripper after grasping~\cite{patel_Robotic_2025,bharadhwaj_Track2Act_2024,yin_Objectcentric_2025,huang_IMAGINATION_2024}, we convert the recovered object trajectory into an end-effector trajectory
    $T_{\mathrm{ee}}^t = T_{\mathrm{obj}}^t T_{\mathrm{obj}\rightarrow \mathrm{ee}}$.
This representation is accurate when the target object is visible, rigid, and consistently tracked. 

\textbf{Hand-centric grounding.}
Object flow becomes unreliable when the target object is heavily occluded by the generated hand, when depth estimates flicker around contact, or when the fitted object motion contains implausibly large rotations.
In such cases, NovaPlan uses the generated human hand as a kinematic prior.
We reconstruct a MANO hand mesh~\cite{DBLP:journals/corr/abs-2201-02610} for each video frame using HaMeR~\cite{pavlakos_Reconstructing_2024}, obtaining a hand trajectory
    $\mathcal{H} = \{ \mathbf{H}^t \mid t=1,\dots,T \}$.

\begin{figure}[t]
    \includegraphics[width=\linewidth]{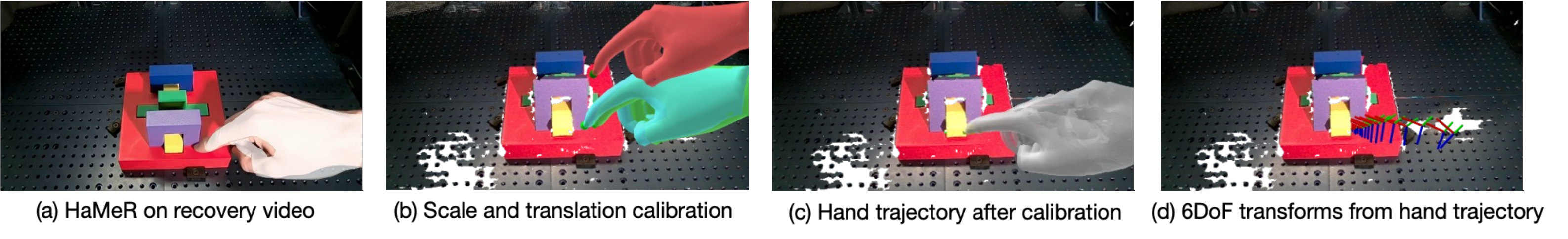}
    \vspace{-15pt}
    \caption{\textbf{Geometric grounding for non-prehensile recovery.} (a) HaMeR~\cite{pavlakos_Reconstructing_2024} predicts the MANO hand mesh on the generated recovery video. (b) \textcolor{red}{Red:} mesh from raw prediction. \textcolor{green}{Green:} mesh after scale calibration. \textcolor{teal}{Cyan:} mesh after additional translation offset enforcing fingertip and object surface contact.  (c) The hand trajectory after calibration. (d) Calibrated trajectory converted into per-frame SE(3) transforms.
    }
\label{fig:recovery_grounding}
\vspace{-15pt}
\end{figure}

Since generated hands often exhibit scale errors and projective drift, the raw hand reconstruction is not directly metric.
We thus apply a lightweight contact-based calibration (detailed in Appendix~\ref{app:hand_flow_calibration}) to the video hand motion: we detect the interaction interval from object-mask motion, align the hand to the object at contact onset, and compensate for residual drift near release.
The calibrated hand motion is converted into a rigid $SE(3)$ trajectory by using the calibrated contact fingertip as the translation reference and a palm frame estimated from hand landmarks as the orientation reference.
The resulting geometric effect is illustrated in \Cref{fig:recovery_grounding}, where the raw hand reconstruction is brought into metric contact with the object and converted into per-frame $SE(3)$ transforms.

\textbf{Flow switching and execution.} NovaPlan first attempts object-centric grounding and switches to hand-centric grounding when the recovered object motion is unstable. 
We reject object flow if tracking fails or if the fitted inter-frame rotation exceeds a predefined threshold, i.e., $\exists t,\; \theta_t > \theta_{\max}$, where $\theta_t$ denotes the axis-angle rotation magnitude of the per-frame fitted object transform and $\theta_{\max}$ is the rotation threshold for triggering the flow switch. 
Appendix~\ref{app:theta_ablation} presents relevant analyses on the design choice of the hand flow and the effect of varying $\theta_{\max}$ on execution performance.
When object flow is rejected, the system switches to hand-centric grounding.
For hand flow, the initial hand--object contact pose guides grasp selection (see Appendix~\ref{app:grasp_selection}), and the robot follows the calibrated hand trajectory after grasping.

For non-prehensile recovery videos, the contact finger is specified by the recovery prompt rather than inferred from fingertip-mask overlap.
We therefore constrain hand-flow calibration to the prompted contact finger, enforcing metric contact between the fingertip and the selected object surface.
This contact-constrained grounding is illustrated in \Cref{fig:recovery_grounding} and described in detail in Appendix~\ref{app:nonprehensile_grounding}.

\vspace{-5pt}
\section{Experiments}
\label{sec:experiments}
\vspace{-3pt}

\method is designed to solve diverse long-horizon manipulation problems in a zero-shot manner.
To evaluate its capabilities, we design a set of long-horizon tasks that assess reasoning, task decomposition, sub-task execution, error diagnosis, and recovery.
We also include tasks from~\cite{li_NovaFlow_2025} to evaluate improvements in short-horizon zero-shot settings.
Finally, we use an assembly task from the Functional Manipulation Benchmark (FMB)~\cite{luo_FMB_2024} to discuss the boundaries of \method under the same zero-shot setup.
Appendix~\ref{app:imp_details} presents implementation details and runtime analysis, where each macro-action computation takes roughly 40\,s end to end.

For baselines, we compare \method against five representative zero-shot methods from distinct architectural paradigms: NovaFlow~\cite{li_NovaFlow_2025} (video-based planning), VLP-style planning~\cite{du_Video_2023} (video-language planning), DreamZero~\cite{ye_DreamZero_2026} (world action model), \textbf{$\pi_{0.5}$}~\cite{intelligence_$p_05$_2025} (VLA model), and MOKA~\cite{liu_MOKA_2024} (VLM-based planning). The VLP-style baseline keeps our video generation and training-free hybrid-flow execution and replaces only the planner, so that the comparison reflects the planning loop rather than choices of backbone models.
NovaFlow, DreamZero, and $\pi_{0.5}$ receive oracle subtask decompositions, isolating grounding and execution for these baselines; MOKA and VLP-style use their own task planners (Appendix~\ref{app:oracles}).
\vspace{-5pt}
\subsection{Long-Horizon Task Solving}
\vspace{-3pt}
\label{sec:exp1}

\begin{figure*}
    \includegraphics[width=\textwidth]{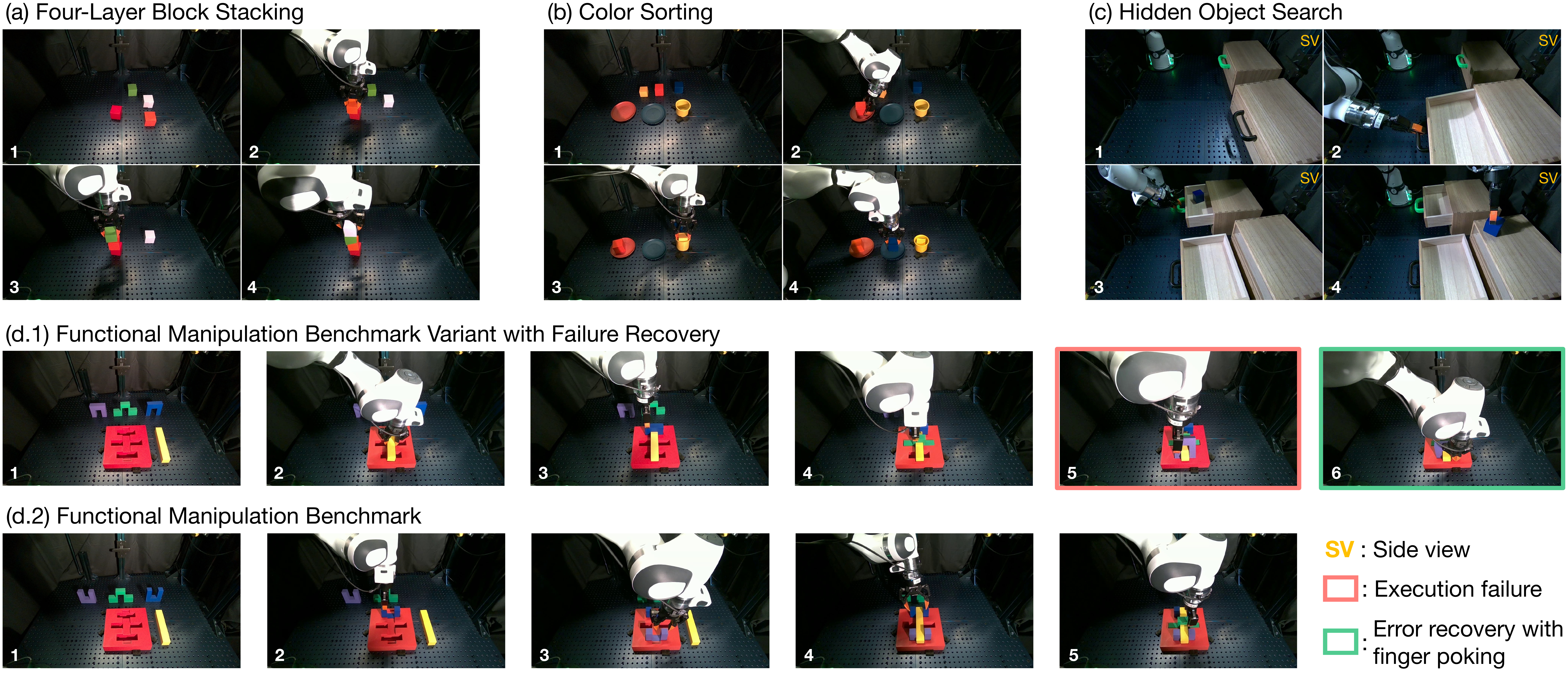}
    \caption{\textbf{Robot Experiments.} (a-c) The long-horizon tasks in \Cref{sec:exp1}, each with three steps. (d) Functional Manipulation Benchmark Multi-Object Multi-Stage Assembly 1 task and its variant in \Cref{sec:exp3}, including one recovery step.}
    \label{fig:exp_rollout}
\vspace{-15pt}
\end{figure*}

\begin{figure*}
    \includegraphics[width=\textwidth]{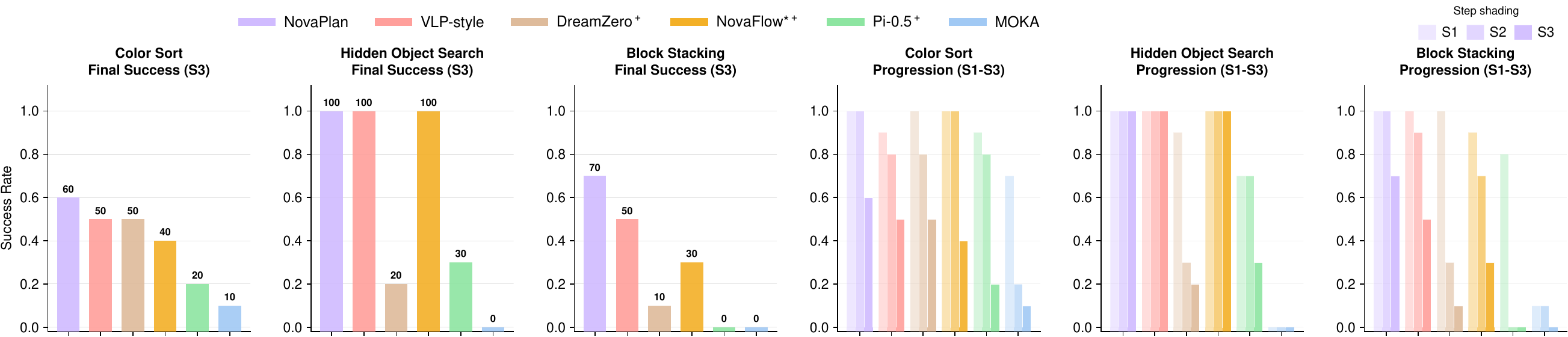}
    \caption{
        \textbf{Long-horizon task performance.}
        (Left) Final success rate of different methods on three long-horizon tasks.
        (Right) Success rate of each method after every step of the three long-horizon tasks, where shading denotes the step index.
        Methods marked with $^+$ (NovaFlow, DreamZero, and $\pi_{0.5}$) receive oracle high-level plans; MOKA and VLP-style generate their own plans.
    }
    \label{fig:long_horizon_result}
    \vspace{-15pt}
\end{figure*}
To test NovaPlan's long-horizon robustness, we evaluate the system's performance on three tasks:

\vspace{-5pt}
\begin{itemize}[leftmargin=0.9em, itemsep=0.1em, topsep=0.1em]
    \item \textbf{Four-layer Block Stacking:} Stack four colored 2-inch blocks into a vertical tower, with each step placing a block onto the increasingly taller stack. This task evaluates spatial reasoning and extracted action accuracy under increasing placement height.
    \item \textbf{Color Sorting:} Sort three blocks into color-matched containers. Beyond semantic reasoning, the yellow block requires precise vertical alignment due to its tight fit to avoid jamming.
    \item \textbf{Hidden Object Search:} Find a target object hidden in one of the two closed drawers and place it on the other drawer's top surface. This exploratory task evaluates conditional planning, requiring a two- or three-step plan depending on which drawer contains the object.
\end{itemize}
\vspace{-5pt}

We conduct ten trials for each task with varying initial setups.
For each step in the long-horizon tasks, we mark the trial as a failure only when the method \textit{cannot recover} from an error. For \textit{Color Sorting}, since sub-tasks are order-independent, we use the step number to represent the number of correctly sorted objects after full execution. We report both overall long-horizon task success rates and task progression based on single-step success rates in~\Cref{fig:long_horizon_result}. For more results on the generated plans and experiment rollout visualizations, please see Appendices~\ref{app:exp_rollout} and~\ref{app:example_plan}.

In \textit{Block Stacking}, performance decreases as the stack grows taller because later placements require more precise spatial reasoning and control: $\pi_{0.5}$ stacks up to two layers but fails beyond that, while NovaFlow reaches the third block with 70\% success but drops to 30\% on the fourth due to object-flow instability. 
By switching to hand flow when object flow becomes unreliable, \method improves execution stability and succeeds in seven out of ten trials. 
In \textit{Color Sorting}, the low-tolerance ``yellow block and cup" step causes all methods to drop in performance; for \method, the remaining failures are mainly caused by inaccurate pose extraction from depth estimation errors, which could be mitigated by stronger depth models or additional camera views. 
In \textit{Hidden Object Search}, success depends on both drawer opening and object retrieval: $\pi_{0.5}$ performs better on drawer opening than retrieval, likely reflecting the task distribution in the DROID dataset~\cite{khazatsky2024droid}, while MOKA fails to find a valid grasp for the horizontal drawer handle. 
Both \method and NovaFlow succeed in all trials, as their sub-skill execution remains reliable across the task sequence. 
VLP-style planning matches \method on reactive search, but degrades on the coupled tasks, as a mismatch between predicted and real states changes later placement geometry, and incurs $2$--$3$ times as many re-planning calls due to mismatch-caused execution failures.
DreamZero degrades sharply after the first action.
Overall, these results show that \method combines stable short-horizon execution with robust switching and recovery, enabling stronger long-horizon manipulation performance.

\begin{figure*}[t!]
    \centering
    \begin{minipage}[t]{0.45\textwidth}
        \centering
        \includegraphics[width=\linewidth]{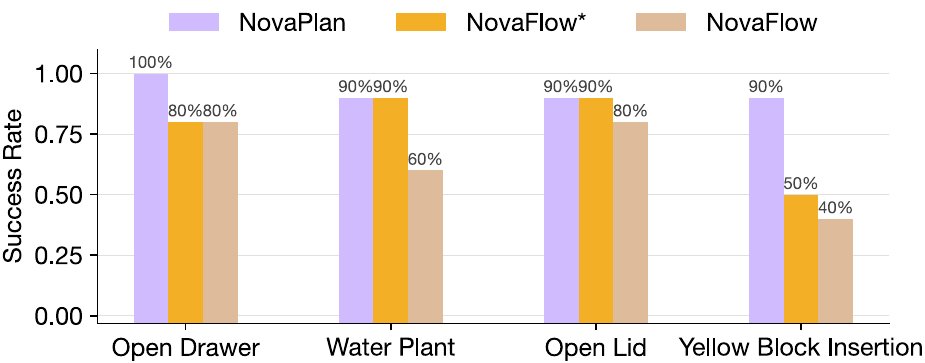}
        \caption{
        \textbf{Single-step task performance.}
        Comparison with NovaFlow* (implemented) and NovaFlow (from \cite{li_NovaFlow_2025}).
        }
        \label{fig:single_step_result}
    \end{minipage}
    \hfill
    \begin{minipage}[t]{0.5\textwidth}
        \centering
        \includegraphics[width=0.95\linewidth]{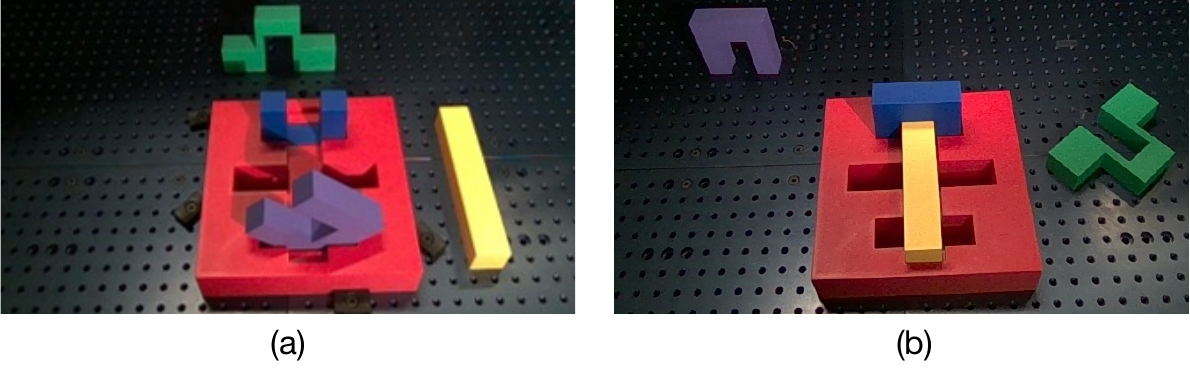}
        \caption{
        \textbf{Failure cases on FMB task} due to (a) video model generating physically implausible motions or (b) the need for object reorientation.
        }
        \label{fig:fmb_failure_recovery}
    \end{minipage}
    \vspace{-15pt}
\end{figure*}

\vspace{-5pt}
\subsection{Short-Horizon Task Solving}
\vspace{-5pt}
\label{sec:exp2}

To evaluate the efficacy of our hybrid flow mechanism, we compare NovaPlan with NovaFlow on its original single-step task suite: \textit{Open Drawer}, \textit{Water Plant}, \textit{Open Lid}, and \textit{Yellow Block Insertion}, using ten trials per task under the same setup. 
We exclude \textit{Hang Mug}, which requires a goal image for plan video generation, and \textit{Cup on Saucer}, whose pick-and-place behavior is well covered by our long-horizon tasks. 
This comparison tests whether hand flow can mitigate self-occlusion failures in object-centric tracking. 
As shown in~\Cref{fig:single_step_result}, \method achieves higher success rates across all tasks, indicating that using the hand as a reference improves stability when object flow is noisy.
\vspace{-5pt}
\subsection{Zero-Shot FMB Solving}
\vspace{-5pt}
\label{sec:exp3}

We use FMB as a case study to probe the capability boundary of \method.
FMB requires the system to compose individual manipulation skills into functionally meaningful long-horizon behaviors, making it a challenging stress test for zero-shot manipulation. 
The task demands long-horizon planning, millimeter-level assembly precision, and irregular object shapes that are often unseen by pre-trained foundation models, with failure modes that cannot be resolved by simple re-grasping. 

Since here \emph{none} of the chosen VLA/VLM baselines can complete even a single step, we focus on the qualitative success and failure mode analysis of \method. 
We evaluate on standard Initial Layout 3 of the FMB Multi-Object Multi-Stage Task Assembly 1, shown in~\Cref{fig:exp_rollout}(d.2), where all parts lie flat on the table and require no reorientation. 
We also test a variant with the U-shaped parts placed upside down, shown in~\Cref{fig:exp_rollout}(d.1), to evaluate execution on a different assembly target.

We identify two main bottlenecks in this setting. 
First, video generation becomes unreliable for irregular assembly objects. 
Only Veo3.1 can generate physically feasible videos for all FMB stages, and even then with a much lower success rate, especially during failure recovery where contact dynamics of the human finger are more complex. In this case, \method keeps regenerating plans until a feasible recovery video is obtained. 
Second, insertion recovery remains difficult because both object displacement and hand motion are small after partial failure. 
As shown in~\Cref{fig:fmb_failure_recovery}, this causes object keypoint flow and hand pose estimates to be dominated by noise, making both object-flow and grasp-mode hand-flow execution unreliable. In these cases,  \method can still ground execution on fingertip motion during non-prehensile pushing, providing a cleaner signal that enables recovery.

\begin{table}[t]
\centering
\caption{\textbf{(a) Component ablations and (b) failure attribution} (\%; 10 trials per task). In (a), each row swaps one module, all else fixed; $^{\ddagger}$ scored by long-horizon task success, others by module-level correctness. In (b), $^{*}$ covers video generation, object/hand segmentation, and recovery reasoning.}
\label{tab:component_results}
\scriptsize
\renewcommand{\arraystretch}{0.95}
\setlength{\tabcolsep}{5pt}
\begin{tabular}{@{}llcc@{\hspace{26pt}}lc@{}}
\toprule
\multicolumn{4}{@{}l}{\textbf{(a) Controlled component ablations}} & \multicolumn{2}{@{\hspace{26pt}}l@{}}{\textbf{(b) Failure attribution by module}} \\
\cmidrule(r{26pt}){1-4} \cmidrule{5-6}
\textbf{Component} & \textbf{Variant} & \textbf{Ours} & \textbf{Ablated} & \textbf{Module} & \textbf{Failure rate} \\
\midrule
Strategic beam search               & Reactive planning     & \textbf{85.0} & 15.0 & 3D object tracking                       & 20.0 \\
Execution re-grounding$^{\ddagger}$ & Reuse planned rollout & \textbf{76.7} & 43.3 & Depth estimation                         & 12.9 \\
Rollout verification                & Without flow image    & \textbf{98.8} & 42.4 & Hand calibration                         & 6.7 \\
Hybrid grounding$^{\ddagger}$       & Object flow only      & \textbf{76.7} & 46.7 & Grasp selection and execution            & 5.9 \\
Hand calibration                    & Raw HaMeR             & \textbf{93.3} & 13.3 & Rollout selection                        & 1.2 \\
State verification                  & Without target frame  & \textbf{98.8} & 92.9 & State verification                       & 1.2 \\
Local recovery$^{\ddagger}$         & Terminate on failure  & \textbf{76.7} & 50.0 & Video generation, segmentation, recovery$^{*}$ & 0.0 \\
\bottomrule
\end{tabular}
\vspace{-15pt}
\end{table}

\subsection{Component Ablations and Failure Attribution}
\label{sec:exp4}
\Cref{tab:component_results}(a) presents the module ablation results, where we replace one module in the pipeline at a time for a variant and observe two patterns. 
First, the pre-execution modules each prevent a potential failure.
Scoring rollouts from frames alone collapses valid selection to 42.4\%, since a goal-like final frame can be reached by motion that never occurs.
Calibration and planning are also meaningful: raw HaMeR output is not metric, and an ordering error cannot be undone once a step blocks access.
Second, the execution-time mechanisms are not incremental refinements.
Removing execution re-grounding, local recovery, or hand flow each costs less than half of the long-horizon success rate.
The switch to hand flow takes up only a modest fraction of steps, but those are the steps that are hard and would otherwise fail a trial (Appendix~\ref{app:theta_ablation}). \Cref{tab:component_results}(b) then shows the breakdown of module failure rates. The reasoning stages are most reliable, and infeasible rollouts are filtered before the robot moves, so \method's bottleneck lies in metric geometry rather than in the planner. Please refer to Appendices~\ref{app:failure_attribution}  and~\ref{app:ablation_protocol} for full details.

\section{Limitations and Conclusion}
\label{sec:limitations}
\textbf{Limitations.}
\method relies on foundation and perception models for vision-language reasoning, video generation, depth estimation, keypoint tracking, and hand pose estimation, and its performance is bounded by their capabilities, with metric geometry the weakest part (\Cref{tab:component_results}(b)).
(1) \textit{Scope}: We evaluate tabletop scenes of rigid object motion under a single fixed camera, and do not address clutter, deformables, or transparent objects.
(2) \textit{Evaluation}: Ten trials per task suffice for real-robot comparison, but not for claims of broad object or scene diversity, and while our prompts are robust to paraphrase, they degrade once the reasoning rubric is removed (Appendix~\ref{app:prompt_sensitivity}).
 (3) \textit{Recovery and safety:} Recovery only chooses between grasping and non-prehensile repair, and video models still fail to generate corrections for many failure modes, leaving reorientation and~\Cref{fig:fmb_failure_recovery}(a) out of reach. Since trajectories come from generated videos, a verified rollout can still be physically wrong, so we give no formal safety guarantee and would not deploy \method unsupervised.
Yet, given its principled interface structure, \method will naturally improve with advances in foundation models and geometric reasoning.

\textbf{Conclusion.}
We introduce \method, a zero-shot, long-horizon manipulation framework coupling closed-loop video-language planning with geometrically grounded execution.
By extracting object keypoints and hand poses from generated videos, \method connects high-level task reasoning with low-level control.
Experiments across diverse long-horizon tasks and the  FMB boundary case study show decent manipulation and error recovery without task-specific demonstrations or training. 

\clearpage
\acknowledgments{This work was fully supported by the Robotics and AI Institute.}

\bibliography{references}

\newpage
\clearpage
\appendix
\textbf{Appendix overview.}
Appendix~\ref{app:imp_details} presents implementation details and discusses runtime.
Appendix~\ref{app:theta_ablation} analyzes the hand-flow design and its sensitivity to the switching threshold.
Appendix~\ref{app:VLM_planning} provides full details of the planning, verification, and recovery procedures, along with the prompts used at each stage.
Appendix~\ref{app:grounding_details} describes the geometric grounding procedure, including flow extraction, hand calibration, and grasp selection.
Appendix~\ref{app:oracles} specifies the baseline setups and oracle information.
Appendix~\ref{app:exp_rollout} provides additional experiment rollout visualizations, while Appendix~\ref{app:example_plan} presents example action plans and video-generation prompts.
Appendix~\ref{app:attribution} reports failure attribution, component ablations, and prompt sensitivity analysis.

\section{Implementation Details and Runtime Analysis}
\label{app:imp_details}
\textbf{Hardware setup.}
We utilize a Franka Research 3 arm equipped with a Robotiq 2F-85 gripper, with RGB and depth images captured by a front-facing RealSense D455 camera fixed on the table. 
All experiments use a joint impedance controller to track the calculated joint reference.

\textbf{Implementation details}. We upgrade several modules from~\cite{li_NovaFlow_2025}: MegaSaM~\cite{li_MegaSaM_2025} to MoGe2~\cite{wang_MoGe2_2025} for metric depth recovery and Grounded-SAM2~\cite{ren_Grounded_2024} to SAM3~\cite{carion_SAM_2025}. 
We use TAPIP3D~\cite{zhang_TAPIP3D_2025} for 3D point tracking, HaMeR~\cite{pavlakos_Reconstructing_2024} for hand tracking, Wan2.2~\cite{wan_Wan_2025, lightx2v} and Veo3.1~\cite{wiedemer_Video_2025} for video generation, and GPT-5.2 as the VLM for all reasoning tasks. We use Wan2.2 and Veo3.1 each to generate $L=8$ rollout videos as candidates for selection.
All perception backbones run with their released default inference settings, and the geometric thresholds are held fixed across every task. The maximum rotation magnitude between frames is set as $\theta_{\max}=45^\circ$ for switching from object to hand flow. We provide an ablation analysis over $\theta_{\max}$ in~\Cref{app:theta_ablation}. 

\textbf{Runtime analysis.} We deploy perception-heavy components, including video generation, monocular depth estimation, and 3D point tracking, on an NVIDIA H100 GPU, while hand-flow extraction is performed on an NVIDIA RTX A6000 workstation. \Cref{tab:runtime} reports the runtime of the main NovaPlan modules for a 41-frame 720P video, including video generation with Wan2.2 or Veo3.1, monocular depth estimation with MoGe2 or with optional CVD-based depth refinement (see \Cref{app:object_flow}), object segmentation with SAM3, object point tracking with TAPIP3D, and hand-flow extraction with HaMeR. With parallelized object- and hand-flow extraction, NovaPlan achieves an end-to-end runtime of approximately 40\,s.

\begin{table}[H]
\vspace{-10pt}
\centering
\caption{Runtime breakdown for major modules with a 41-frame 720P video (seconds).}
\label{tab:runtime}
\small
\begin{tabular}{@{}lccccr@{}}
\toprule
\textbf{Model} & \textbf{Wan / Veo} & \textbf{MoGe2 / MoGe2+CVD} & \textbf{SAM3} & \textbf{TAPIP3D} & \textbf{HaMeR} \\ 
\midrule
\textbf{Time}  & 30 & 3 / 90 & 3.5 & 3.5 & 0.8/frame \\ 
\bottomrule
\end{tabular}
\vspace{-5pt}
\end{table}

\textbf{End-to-end wall-clock time.}
Aggregating \Cref{tab:runtime} at the task level, video generation dominates planning at roughly 30\,s of the $\sim$40\,s spent per macro-action, and including network latency and robot execution, each macro-action takes about 90\,s.
A reactive two- to three-step task therefore takes $\sim$3--4.5\,min without recovery and $\sim$4.5--6\,min with one recovery cycle. Most runs need no recovery, and when they do it adds one step cycle to the total time.
Three-step coupled tasks additionally pay $\sim$2.8\,min for strategic beam search before execution begins, giving $\sim$7--9\,min in total.
Since planning cost is incurred once per macro-action, end-to-end time grows roughly linearly with the task horizon.

\section{Analysis of Hand-Flow Grounding}
\label{app:theta_ablation}
\textbf{How often hand flow is used.}
Hand flow acts as a fallback, so its value depends on how often it is actually invoked.
Across the long-horizon experiments, hand flow is selected for $7/30$ block-stacking actions, $6/30$ color-sorting actions, and $1/25$ hidden-object-search actions.
The overall utilization ratio is therefore modest.
However, the switches concentrate on the steps that are the hardest and would otherwise terminate a trial, e.g., placing the fourth block or inserting the tightly fitting yellow block.
This is consistent with the hybrid-grounding ablation in \Cref{tab:component_results}(a), where disabling the switch lowers long-horizon success from 76.7\% to 46.7\%.

\textbf{Why a human hand.}
We use a human hand, rather than a robot end-effector, as the source morphology for video generation for two reasons.
First, current video generation models produce physically plausible human manipulation far more reliably than robot arm motion, since human-related data are more common than robot data in their training distribution.
Second, the hand stays partially visible and informative even when the manipulated object is occluded, which is exactly the regime in which object flow becomes unreliable.
We emphasize that \method does not imitate human joint angles, so only the calibrated task-space $SE(3)$ motion of the designated contact fingertip is retargeted to the gripper, as described in \Cref{app:hand_flow_calibration}.

\textbf{Sensitivity to the switching threshold.}
We analyze the sensitivity of the hand-flow switching threshold $\theta_{\max}$, which controls when \method will switch from object-flow-based execution to hand-flow-based execution. In our main experiments, we use $\theta_{\max}=45^\circ$ as a simple heuristic default corresponding to a visually large inter-frame rotation, rather than as a heavily tuned hyperparameter.

To evaluate whether performance depends sensitively on this choice, we conduct a sensitivity analysis over $\theta_{\max}\in\{20^\circ,30^\circ,45^\circ,60^\circ\}$, along with an  object-flow-only variant  ($\theta_{\max}=\infty$) that disables hand-flow switching. We evaluate on the two most failure-prone steps in the long-horizon experiments and report single-step success rates over 10 trials per setting.

Across both tasks, enabling a switch to hand flow consistently outperforms object-flow-only execution. Thresholds in the $20^\circ$--$45^\circ$ range perform similarly well, suggesting that \method is not highly sensitive to the exact value of $\theta_{\max}$. Performance degrades slightly at $60^\circ$, likely because the system is too tolerant to switch away from unreliable object flow. These results support our key design choice of allowing execution to switch from object flow to hand flow when object tracking becomes unreliable, while indicating that $\theta_{\max}$ itself is not a fragile threshold.

\begin{table}[H]
\vspace{-5pt}
\centering
\caption{Ablation study of the hand-flow switching threshold $\theta_{\max}$ on single-step success rates.}
\label{tab:theta_ablation}
\small
\begin{tabular}{lccccc}
\toprule
$\bm{\theta_{\max}}$ 
& \textbf{20$\bm{^\circ}$} 
& \textbf{30$\bm{^\circ}$} 
& \textbf{45$\bm{^\circ}$} 
& \textbf{60$\bm{^\circ}$} 
& \textbf{$\bm{\infty}$ (Object-flow Only)} \\
\midrule
\textbf{4-layer Block Stacking: 4th Block} & 70\% & 60\% & 70\% & 60\% & 40\% \\
\textbf{Color Sorting: Yellow Block} & 70\% & 60\% & 60\% & 40\% & 20\% \\
\bottomrule
\end{tabular}
\vspace{-5pt}
\end{table}

\section{Vision-Language Planning Implementation Details}
\label{app:VLM_planning}
To ensure robust long-horizon reasoning, our high-level planner operates within a physics-compliant reasoning framework established via structured prompt conditioning. Rather than relying on unconstrained generation, this approach restricts the model's output space to adhere to environmental affordances and causal dependencies. The following sections detail the specific reasoning strategies governing our VLM prompting. 

\subsection{Constraint-Aware Task Decomposition and Action Proposal}
\label{app:high_level_action_proposal}
The high-level planner takes as input the task goal, the latest scene observation, and the action history. It decomposes the goal into a sequence of macro-actions. To reduce physically invalid proposals and redundant moves, we use a structured prompt that enforces basic physical constraints and ordering rules.

\subsubsection{Dynamic Context and Prompt Structure}
\label{app:prompt_action_proposal}

The VLM action proposal is conditioned on the current observation, the task goal, and the full action history. It also receives a small amount of step-dependent context: the remaining planning horizon $T_{remain}$ and an optional \emph{constraint context} carried over from the previous step (when available). The prompt encodes three types of guidance.

\begin{itemize}
    \item \textbf{Horizon-dependent urgency.} We add a short note based on $T_{remain}$. If $T_{remain}\le 2$, we append \textit{``Time is critical. Actions must be aggressive.''} Otherwise, we append \textit{``You have time. Prioritize precision and safety.''}
    \item \textbf{State awareness to avoid redundant moves.} The model is instructed to treat objects that are already in their final configuration as \emph{completed} and to avoid moving them, while treating misplaced objects that overlap target zones as potential obstructions.
    \item \textbf{Ordering constraints.} When two objects interact, the model is instructed to test both orders. If executing $A$ then $B$ is feasible but $B$ then $A$ blocks access, then $A$ is enforced as the prerequisite, and all proposed actions must start with $A$.
\end{itemize}

\begin{promptbox}[Prompt for Task Decomposition and Action Proposal]
\label{planner_prompt}
Analyze the scene and history. Propose EXACTLY \{num\_actions\} distinct narrative macro-actions.

\textbf{Goal:} \{goal\} \\
\textbf{Action History:} \{prev\_act\_str\} \\
\{urgency\_note\} \\
\textbf{Constraint Context:} \{constraint\_context\}

\textbf{Critical visual analysis (intersections \& hierarchy)}
\begin{enumerate}
    \item \textbf{Trace target zones:} Visually trace the target locations (slots, containers, stacking areas) for \emph{all} objects.
    \item \textbf{Detect overlaps:} Do any target zones visually cross or intersect?
    \item \textbf{Determine hierarchy (access \& occlusion):}
    \begin{itemize}
        \item \textbf{Occlusion principle:} If placing object $B$ visually ``covers'' or ``bridges'' the access path to object $A$'s target, then object $A$ must be placed first.
        \item \textbf{Constrained access:} Placing objects in \emph{deep} or \emph{central} zones first is safer than placing peripheral objects that may narrow the workspace.
        \item \textbf{Generic rule:} Place objects that sit \emph{lowest} or \emph{deepest} in the assembly structure first.
    \end{itemize}
\end{enumerate}

\textbf{State verification (goal state vs.\ obstruction)}
\begin{itemize}
    \item \textbf{Analyze objects in the workspace:} Distinguish objects that are \emph{correctly placed} from those that are \emph{clutter/obstacles}.
    \item \textbf{Correctly placed:} If an object is fully aligned with its target (e.g., inside its slot, neatly stacked, or in its final position), treat it as a completed step, not an obstruction. It usually does not need to be moved.
    \item \textbf{Misplaced/loose:} If an object lies haphazardly across target zones or rests on top of other destinations without being properly placed, it is an obstruction that might need to be moved.
    \item \textbf{Correction:} Do not propose removing objects that are already in their correct final state unless they clearly block a necessary path.
\end{itemize}

\textbf{Mental simulation (``order of operations'' test)}

Compare potential sequences for interacting objects ($A$ and $B$):
\begin{enumerate}
    \item \textbf{Sequence 1 ($A$ then $B$):} ``If I place $A$ first, can I still place $B$?''
    \item \textbf{Sequence 2 ($B$ then $A$):} ``If I place $B$ first, can I still place $A$?''
\end{enumerate}

\textbf{Critical conclusion}
\begin{itemize}
    \item If ($A$ then $B$) works but ($B$ then $A$) fails because $B$ blocks $A$'s path/slot, then $A$ is the prerequisite. You must propose $A$ first.
    \item \textbf{Generic indicator:} The prerequisite object often lies ``under,'' ``behind,''  or ``inside'' the other object's zone, or its insertion path is crossed by the other object.
    \item \textbf{Risk assessment:} If paths cross or zones overlap, assume strict ordering is required until proven otherwise.
\end{itemize}

\textbf{Selection filter}
\begin{itemize}
    \item Only propose actions that pass the ``order of operations'' test.
    \item If multiple valid actions exist (independent sub-tasks), propose diverse options.
    \item \textbf{Mandatory filter:} If you identify a prerequisite object (e.g., $A$ must be first), then every single one of the \{num\_actions\} proposals must start with $A$.
    \item Do not offer options that violate the prerequisite. If $A$ is the bottleneck, diversity should come from \emph{how} you handle $A$ (e.g., different grasp or approach), not from proposing $B$ or $C$.
\end{itemize}

\textbf{Step 1: phase identification} \\
Where are we? (approaching $\rightarrow$ interacting $\rightarrow$ transporting $\rightarrow$ finishing).

\textbf{Step 2: apply constraints (critical)}
\begin{enumerate}
    \item \textbf{No telekinesis (active voice):} Objects never move on their own.
    \begin{itemize}
        \item \textbf{Bad:} ``The pot moves to the drawer.''
        \item \textbf{Good:} ``Grasp the pot and move it.''
    \end{itemize}
    \item \textbf{Abstraction level:}
    \begin{itemize}
        \item \textbf{Bad:} ``Move forward 10\,cm.''
        \item \textbf{Good:} ``Reach down and grasp the handle.''
    \end{itemize}
    \item \textbf{Single hand only:} The robot has one right hand. Do not use bimanual actions.
    \item \textbf{Pacing \& termination:}
    \begin{itemize}
        \item Do not complete the task instantly if the object is far; approach first.
        \item If done: output \texttt{FINISH: Hold position}.
    \end{itemize}
    \item \textbf{Action description format:}
    \begin{itemize}
        \item Do not mention ``the robot'' or ``the robot right hand.''
        \item Start directly with the verb.
        \item Be definitive; do not use ``/'' or ``or'' inside an action.
        \item \textbf{Bad:} ``Grasp the block from the top/side.''
        \item \textbf{Good:} ``Grasp the block from the top.''
    \end{itemize}
\end{enumerate}

\textbf{Step 3: generate \{num\_actions\} strategies} \\
For each action, identify the \textbf{primary object} being manipulated or interacted with. This object will be tracked visually during video generation.

\textbf{Output format (JSON only). You must provide EXACTLY \{num\_actions\} actions:}\\
\{\\
\hspace*{1em}"phase": "Current Phase Analysis...",\\
\hspace*{1em}"dependency\_analysis": "Explain the 'Future Block' test results.",\\
\hspace*{1em}"valid\_objects": ["List of objects that are safe to manipulate NOW"],\\
\hspace*{1em}"proposals": [\\
\hspace*{2em}\{"action": "Action description 1", "track\_object": "object being manipulated (e.g., drawer handle, cup, lid)", "reasoning": "Why this is valid now?"\},\\
\hspace*{2em}\{"action": "Action description 2", "track\_object": "object being manipulated", "reasoning": "Why this is valid now?"\},\\
\hspace*{2em}...\\
\hspace*{1em}]\\
\}\\

Important: the \texttt{proposals} array must contain EXACTLY \{num\_actions\} items. Each item must include both \texttt{action} and \texttt{track\_object}.
\end{promptbox}
We post-process the returned actions to remove boilerplate prefixes (e.g., ``the robot \ldots''), enforce single-hand phrasing, and ensure that exactly the desired number of actions is produced. In greedy mode (see~\Cref{sec:video_planning}), we set $N_c=1$ and propose the next viable action based on the latest observation. In strategic mode, we perform tree search and use the video generation model as an object dynamics verifier to select a full action plan in advance; details are given in~\Cref{sec:tree_search}. \Cref{app:example_plan} shows example multi-step plans for all long-horizon tasks in our experiments.

\subsubsection{Full Action Plan Generation in Strategic Mode}
\label{sec:tree_search}
As described in~\Cref{sec:video_planning}, for coupled tasks we use a strategic beam search to generate a single feasible plan over the full horizon before execution. At each step in strategic mode, we ask the VLM to propose $N_c=2$ action candidates per step and sample $L=4$ video rollouts per action candidate. We then select the top $A=2$ beams whose rollouts receive the highest rankings and use them to initialize the next step. This process continues until the final step, where the best beam is selected to form the full-horizon plan. The full procedure is summarized in~\Cref{alg:vlp_beam_search}.

\begin{figure}[t]
    \centering
    \includegraphics[width=\linewidth]{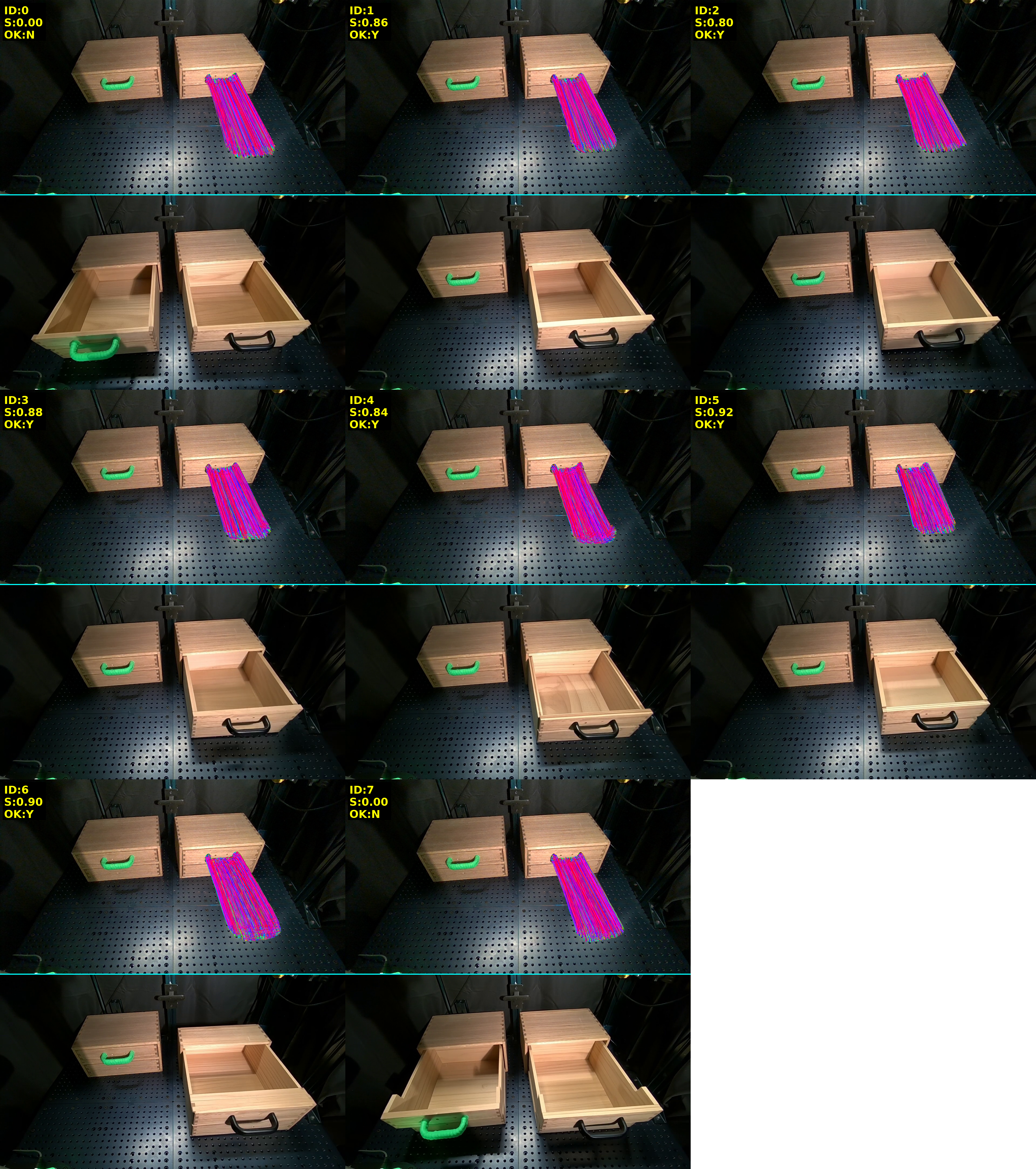}
\caption{\textbf{Example grid image for video rollout selection.} The grid contains eight candidate video rollouts for the hidden object search task at the first step (``open the right drawer'').}
    \label{grid_image}
    \vspace{-10pt}
\end{figure}

\begin{algorithm*}
    \caption{Strategic Beam Search for Full Task Plan Generation}
\label{alg:vlp_beam_search}
\begin{algorithmic}[1]
\Require start frame $I_0$, goal $g$, horizon $H$
\Require candidate set size $A$, actions per candidate $N_c$, rollouts per action $L$
\Require video generation model $\mathcal{G}$, VLM $\mathcal{V}$
\Ensure best candidate $b^\star$

\vspace{0.25em}
\Statex \textbf{candidate state.} each candidate (beam) $b$ stores:
\Statex \hspace{1.0em} current frame $I_b$; action history $\mathcal{H}_b$; metadata list $\mathcal{M}_b$; score $s_b$.
\Statex \textbf{metadata.} each entry $m \in \mathcal{M}_b$ is a per-step dictionary used for state carryover, e.g.,
\Statex \hspace{1.0em} \{action: $a$, track\_object: $o$, candidate\_id: $id$, VLM\_context: $\xi$\}.
\Statex \textbf{context carryover.} $\textsc{GetContext}(\mathcal{M}_b)$ returns the most recent VLM\_context (or $\emptyset$) to enforce constraints consistently across steps.
\vspace{0.25em}

\State $\mathcal{B} \gets \{\, (I_0, \emptyset, \emptyset, 0)\,\}$

\For{$t \gets 1$ \textbf{to} $H$}
    \State $\mathcal{C} \gets \emptyset$ \Comment{all rollout candidates at step $t$}
    \State $id \gets 0$

    \ForAll{$b=(I_b,\mathcal{H}_b,\mathcal{M}_b,s_b) \in \mathcal{B}$}
        \State $h \gets \textsc{FormatHistory}(\mathcal{H}_b)$
        \State $c \gets \textsc{GetContext}(\mathcal{M}_b)$
        \State $\Pi \gets \mathcal{V}.\textsc{ProposeActions}(I_b, g, N_c, h, H-t+1, c)$
        \Comment{$\Pi$ is a list of proposals $\{(a_i,o_i,\xi_i)\}_{i=1}^{N_c}$}

        \For{$i \gets 1$ \textbf{to} $|\Pi|$}
            \State $a \gets \Pi[i].\text{action}$;\;\; $o \gets \Pi[i].\text{track\_object}$;\;\; $\xi \gets \Pi[i].\text{VLM\_context}$
            \State $\{V^{(1)},\dots,V^{(L)}\} \gets \mathcal{G}.\textsc{GenerateRollouts}(I_b, a, L)$
            \Comment{$V^{(k)}$ is the $k^{th}$ generated video}

            \For{$l \gets 1$ \textbf{to} $L$}
                \State $I' \gets \textsc{LastFrame}(V^{(l)})$
                \State $F \gets \textsc{Flow}(V^{(l)}, o)$ 
                \State $\mathcal{C} \gets \mathcal{C} \cup \{(id, b, a, o, \xi, V^{(l)}, F, I')\}$
                \State $id \gets id + 1$
            \EndFor
        \EndFor
    \EndFor

    \If{$\mathcal{C}=\emptyset$}
        \State \textbf{break}
    \EndIf

    \State $\mathcal{R} \gets \mathcal{V}.\textsc{RankRolloutsBatch}(g, \mathcal{C}, \text{top\_n}=A)$
    \Comment{VLM scores candidates using a stitched grid of (flow, final-frame) tiles}

    \State initialize default score map $s(\cdot) \gets s_{min}$ for all candidate ids
    \ForAll{$r \in \mathcal{R}$}
        \State $s(r.\text{candidate\_id}) \gets r.\text{score}$
    \EndFor

    \State $\mathcal{B}_{next} \gets \emptyset$
    \ForAll{$(id, b, a, o, \xi, V, F, I') \in \mathcal{C}$}
        \State $\mathcal{H}' \gets \mathcal{H}_b \, \cup \{a\}$
        \State $m' \gets \{\,\text{action}:a,\;\text{track\_object}:o,\;\text{candidate\_id}:id,\;\text{VLM\_context}:\xi\,\}$
        \State $\mathcal{M}' \gets \mathcal{M}_b \, \cup \{m'\}$
        \State $b' \gets (I', \mathcal{H}', \mathcal{M}', s(id))$
        \State $\mathcal{B}_{next} \gets \mathcal{B}_{next} \cup \{b'\}$
    \EndFor

    \State $\mathcal{B} \gets \textsc{TopB}(\mathcal{B}_{next}, A)$ \Comment{sort by candidate score and keep best $A$}
\EndFor

\State $b^\star \gets \arg\max_{b \in \mathcal{B}} s_b$
\State \Return $b^\star$
\end{algorithmic}
\end{algorithm*}

\subsection{Physics-Based Video Rollout Selection}
\label{sec:video_selection}

To select a rollout for execution, we use a VLM as an automatic evaluator and apply a hierarchical scoring rubric over the generated candidates. In our experiments, for each proposed action at each step during execution, we use Wan2.2 and Veo3.1 each to generate $L=8$ rollout videos as candidates for selection.

To reduce evaluation cost, we batch-score candidates by stitching them into a single grid image. Each tile corresponds to one candidate rollout and is labeled with its original candidate index (ID).

Concretely, each tile contains (i) a flow image (2D object motion flow overlaid on the start frame from CoTracker3~\cite{karaev_CoTracker3_2024}) and (ii) the last frame of the generated video, stacked vertically and separated by a cyan divider. Given the grid and the associated action text for each ID, the VLM assigns a score to every candidate and follows a strict hierarchy of checks, stopping at the first violated constraint:

\begin{itemize}
    \item \textbf{Target check.} Did the correct object move? If the wrong object moves or there is no motion, the score is $0.0$.
    \item \textbf{Physics check.} Is the motion physically plausible (no melting, deformation, teleportation, or objects appearing from nowhere)? If violated, the score is at most $0.1$.
    \item \textbf{Motion check.} Does the motion direction and trajectory in the flow visualization match the action language instruction? If violated, the score is at most $0.2$.
    \item \textbf{Result check.} Does the final frame match the intended outcome and make progress toward the goal? If satisfied, the score lies in $[0.3, 1.0]$.
\end{itemize}

In \Cref{grid_image} we show an example of the grid image, where each tile is labeled with the returned score and success flag.

\begin{promptbox}[Prompt for Video Rollout Selection]
Rank \{N\} robot rollouts. 

\textbf{Goal:} ``\{goal\}''

\textbf{IMAGE:} Grid with \{N\} tiles. Each tile has TOP=motion flow overlay on initial frame, CYAN LINE=divider, BOTTOM=final frame. Yellow ``ID: k'' label on each tile.

\textbf{Actions:} \\
\{ID 0: ``...''\} \\
\{ID 1: ``...''\} \\
\ldots

SCORING RULES (Strict Hierarchical Check - apply in order, stop at first failure):

1. \textbf{Target Check:} Did the CORRECT object move? If WRONG object moved or NO motion at all $\rightarrow$ score=0.0

2. \textbf{Physics Check:} Is physics realistic? If object MELTS/DEFORMS/TELEPORTS or NEW objects appear from nowhere $\rightarrow$ score$\le$0.1

3. \textbf{Motion Check:} Look at TOP (flow overlay on initial frame). Does the motion direction/trajectory match what the ACTION describes?
   \begin{itemize}
      \item NO (wrong direction/drifted/stuck/incomplete path) $\rightarrow$ score$\le$0.2, ok=false
      \item YES (motion trajectory matches action description) $\rightarrow$ continue
   \end{itemize}

4. \textbf{Result Check:} Look at BOTTOM (final frame, below cyan line). Does it show the expected outcome of the ACTION?
   \begin{itemize}
      \item NO (object deformed/new objects appeared/wrong final position/action not completed) $\rightarrow$ score$\le$0.2, ok=false
      \item YES (final state matches what action intended, progresses toward goal) $\rightarrow$ score 0.3--1.0, ok=true
   \end{itemize}

\textbf{OUTPUT:} JSON only. Score ALL candidates. \\
\{ \\
\hspace*{1em}``rankings'': [ \\
\hspace*{2em}\{``candidate\_id'': 0, ``success'': true, ``score'': 0.9, ``reason'': ``perfect insertion''\}, \\
\hspace*{2em}\{``candidate\_id'': 1, ``success'': false, ``score'': 0.1, ``reason'': ``wrong object moved''\} \\
\hspace*{1em}] \\
\}

Key: candidate\_id=ID (0-based index matching the grid labels), success=boolean, score=float(0--1), reason=string rationale
\end{promptbox}

\subsection{Closed-Loop State Verification}
\label{app:recovery_impl}

After each executed action, \shortname verifies the outcome with a VLM-based critic. Unlike open-loop pipelines that assume success, we explicitly evaluate the observed state transition to detect action-level failures, such as grasp slippage, partial insertions, or acting on the wrong object.

\subsubsection{State Verification}
\label{app:state_verification}
Given an intended action and goal, the critic compares three images:
\begin{itemize}
    \item \textbf{Start state ($I_t$):} the observation before execution.
    \item \textbf{Current state ($I_{t+1}$):} the observation after the robot stops.
    \item \textbf{Target state ($I_{target}$):} the last frame of the generated video plan.
\end{itemize}
The planned video provides a concrete visual reference for the expected outcome. The critic checks whether the change observed in the real world ($I_t \rightarrow I_{t+1}$) is consistent with the intended change in the video plan ($I_t \rightarrow I_{target}$). This comparison distinguishes \emph{no-change} outcomes (the action did not take effect) from \emph{incorrect-change} outcomes (a change occurred but the result is wrong), such as moving the wrong object or stopping before the action is completed.

\begin{promptbox}[Prompt for State Verification]
You are a Robotic Systems Verifier. Your task is to determine whether the CURRENT STATE (Image 2) satisfies the physical requirements of the intended ACTION, given the START STATE (Image 1).

\textbf{INTENDED ACTION:} \{action\} \\
\textbf{GOAL SPECIFICATION:} \{goal\}

You are provided with three images: \\
\textbf{Image 1 (START STATE):} state before the action. \\
\textbf{Image 2 (CURRENT STATE):} state after attempting the action. \\
\textbf{Image 3 (TARGET STATE):} expected outcome from the video plan.

\textbf{Verification logic:}
\begin{enumerate}
    \item \textbf{Analyze the action.} List the physical constraints implied by ``\{action\}'' (e.g., \emph{insert} $\Rightarrow$ inside a cavity and flush; \emph{place} $\Rightarrow$ resting on a surface; \emph{grasp} $\Rightarrow$ firm contact).
    \item \textbf{Compare Image 1 vs.\ Image 2.} What changed? Did the intended object move as expected?
    \item \textbf{Inspect Image 2.} Are the action-specific physical constraints satisfied in the current state?
    \item \textbf{Compare Image 2 vs.\ Image 3.} Using Image 3 as a visual reference, is the object state close enough to the target outcome?
\end{enumerate}

\textbf{OUTPUT (JSON only):} \\
\{ ``success'': boolean, ``reason'': string \}

The \texttt{reason} should identify which physical constraint is (or is not) satisfied, and whether the change from Image 1 to Image 2 matches the intended change from Image 1 to Image 3.
\end{promptbox}

\subsubsection{Recovery Action Proposal and Video Generation}
\label{app: recovery_proposal}

A core capability of \shortname is recovery from runtime errors. When the verifier returns \texttt{"success": false}, the system enters a corrective planning loop. We use hierarchical prompting that prioritizes local corrections over full resets when the remaining discrepancy is small and the goal remains achievable. This design reduces unnecessary motion and limits error accumulation that can arise from repeatedly replanning full pick-and-place sequences.

\textbf{Recovery mode selection.}  The VLM first serves as a decision-maker. It compares the current observation with the target state $I_{target}$ and selects a recovery strategy $S\in\{\mathrm{grasp},\mathrm{non\mbox{-}prehensile}\}$. If grasp recovery is selected, the system proceeds with the standard planning pipeline (i.e., a normal non-recovery step). Otherwise, it invokes the non-prehensile recovery procedure described below.

\textbf{Spatial grounding and annotation.}  When non-prehensile recovery is selected, the system grounds the abstract corrective plan in the scene geometry. We query the VLM to identify a single contact region where a small external interaction (e.g., a poke or nudge) can move the manipulated object toward the target configuration. To bridge the semantic planner and the video generation model, the VLM produces two outputs:
\begin{itemize}
\item \textbf{Visual annotation.} The VLM edits the latest observation image by overlaying a visual marker (e.g., a red star) at the selected contact point, yielding $I_{t+1, anno}$. See~\Cref{red_star} for an example in which the VLM marks the contact point between the purple block and a human index finger with a red star for a corrective poke.

\begin{figure}[t]
\centering
\includegraphics[width=0.9\linewidth]{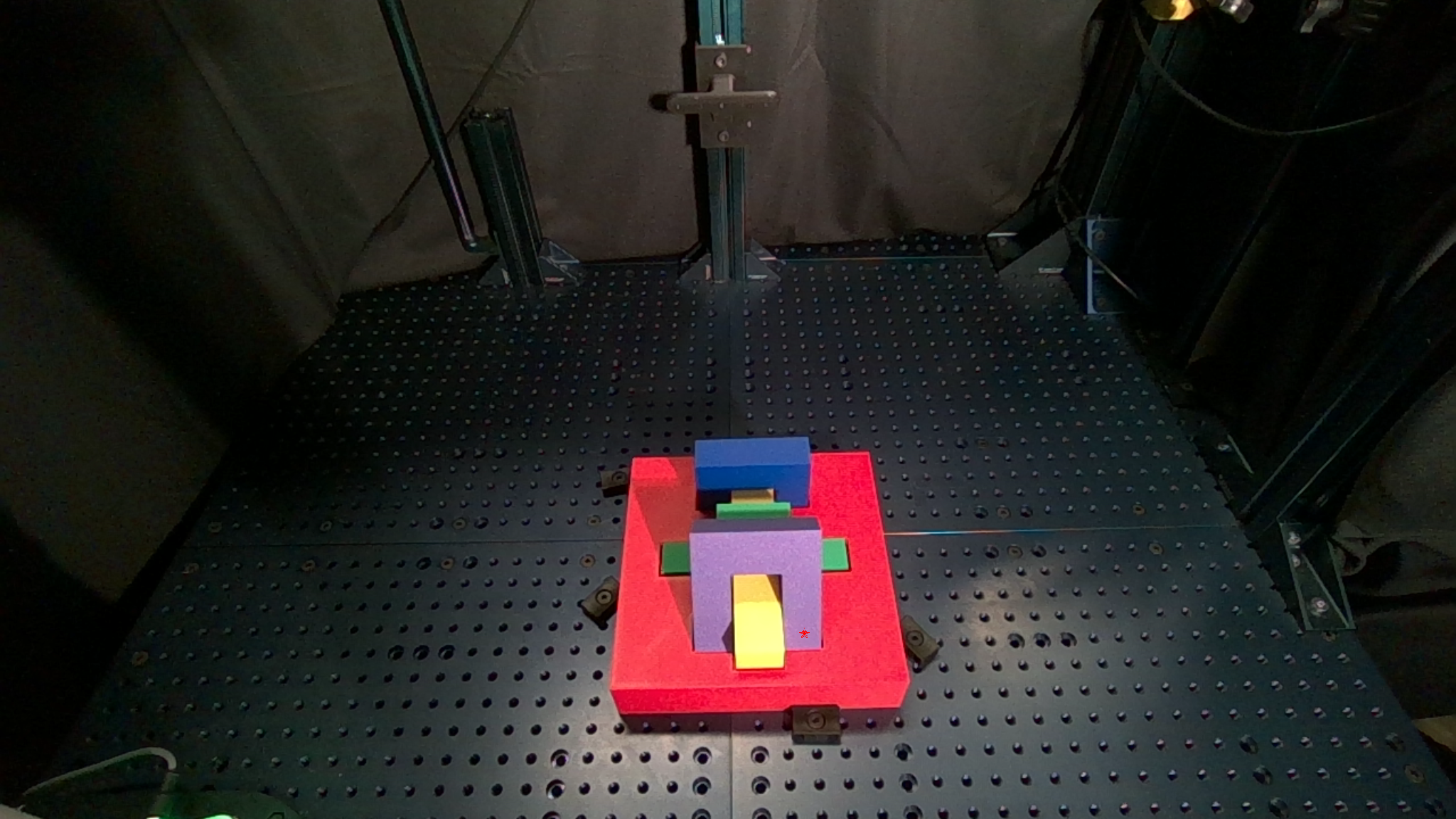}
\caption{\textbf{Example annotated start image for recovery video generation}. The VLM marks the contact point between the purple block and the human index finger with a red star on the lower right part of the purple block, indicating where the corrective poke should occur.}
\label{red_star}
\vspace{-1em}
\end{figure}

\item \textbf{Textual guidance.}  The VLM generates a detailed prompt $P$ that specifies the desired correction video, including the scene setup, the finger contact point, the hand motion trajectory, and the intended final state.
\end{itemize}

\begin{promptbox}[Prompt for Recovery Action Proposal and Video Generation]
You are a Recovery-Policy Controller for a manipulation task. You must (i) select a recovery mode and (ii) if non-prehensile recovery is chosen, spatially ground a single contact point, return an annotated image, and synthesize a structured prompt for video generation.

\textbf{INPUTS (provided):}
\begin{itemize}[leftmargin=1.2em, itemsep=0.2em, topsep=0.2em]
  \item Current observation and image to annotate: \textbf{Image 1}
  \item Goal / target image: \textbf{Image 2}
  \item Goal text: \{goal\_text\}
\end{itemize}

\textbf{TASK 1: Recovery Mode Selection.}

Compare \textbf{Image 1} with \textbf{Image 2} and choose
\[
S \in \{\texttt{grasp},\; \texttt{non\_prehensile}\}.
\]
Choose \texttt{non\_prehensile} if a small perturbation (e.g., a gentle poke/nudge) can plausibly correct the remaining discrepancy; otherwise choose \texttt{grasp}. Provide a brief justification grounded in visible evidence.

\textbf{TASK 2: Spatial Grounding + Annotation (only if $S=\texttt{non\_prehensile}$).}

Identify \emph{one} manipulated object and \emph{one} contact point that will nudge the object toward \textbf{Image 2}.
\begin{itemize}[leftmargin=1.2em, itemsep=0.2em, topsep=0.2em]
  \item Specify: \texttt{object\_name}, \texttt{contact\_surface}, \texttt{contact\_location}.
  \item Define a \textbf{single anchor} contact point (no alternatives).
  \item Overlay a \textbf{small solid red star} centered exactly at the contact point on \textbf{Image 1}, producing the \textbf{annotated Image 1}.
  \item Return the \textbf{annotated Image 1} in the response as a base64-encoded PNG (no surrounding text).
\end{itemize}

\textbf{TASK 3: Textual Guidance Prompt $P$ (only if $S$ is \texttt{non\_prehensile}).}

Write a detailed prompt $P$ for a first-last-frame conditioned video generator, using:
\[
\text{START}=\textbf{annotated Image 1},\qquad \text{GOAL}=\textbf{Image 2}.
\]
$P$ must be highly structured with labeled sections and must include:
\begin{itemize}\item Global constraints (single continuous shot, fixed camera, no cuts; last frame matches goal exactly).
  \item Scene description (what is present; what must remain still / never be touched).
  \item Framing constraints (camera view, margins, full visibility requirements).
  \item Contact target (single anchor; explicitly tie to the red star center).
  \item Hard spatial constraints (where the hand may move; what must remain in frame).
  \item Hand appearance + stability constraints (realistic proportions; no warping; stable pose).
  \item Hand approach constraints (straight-line approach; height constraints; no arcs).
  \item Continuous-contact rule (no sliding; no switching surfaces; no second contact point).
  \item Action description (direction and components; intended rotation/settling toward goal).
  \item End behavior (pause, retract, exit; no second push; realistic contact physics; silent).
\end{itemize}
Use unambiguous language (\textit{must/only/never}), include \textbf{exactly one} contact point and \textbf{exactly one} push, do not introduce new objects, and keep all untouched objects unchanged.

\textbf{OUTPUT (strict JSON only).} Return \emph{only} valid JSON with exactly this schema:
\begin{verbatim}
{
  "recovery_mode": "grasp" | "non_prehensile",
  "mode_justification": {
    "discrepancy_summary": "string (1–3 sentences)",
    "why_this_mode": "string (1–3 sentences)"
  },
  "annotation": {
    "enabled": boolean,
    "object_name": "string" | null,
    "contact_point_definition": "string" | null,
    "edit_spec": {
      "marker": "solid_red_star",
      "anchor": "center_of_star_is_contact_point",
      "placement": "string"
    },
    "annotated_image_png_base64": "string" | null
  },
  "prompt_P": {
    "enabled": boolean,
    "title": "string",
    "text": "string (full structured prompt P)"
  }
}
\end{verbatim}

If \texttt{recovery\_mode}=\texttt{grasp}, set \texttt{annotation.enabled=false} and \texttt{prompt\_P.enabled=false}, and set image/prompt fields to \texttt{null}.
\end{promptbox}

\textbf{Conditioned video generation.}  The final recovery video is generated by a video generation model. We use Veo3.1 to generate the recovery video in the First-Last-Frame setting. Specifically, the model generates a recovery video $V$ conditioned on the annotated start image $I_{t+1, anno}$, the proposed correction prompt $P$, and the target image $I_{target}$ as the required last frame:
\begin{equation}
V = \text{VideoModel}(I_{t+1, anno}, P, I_{target})
\end{equation}
Using $I_{t+1, anno}$ rather than the raw observation constrains the video model to the specific contact dynamics selected by the VLM. This yields a targeted non-prehensile correction (e.g., a single poke) that steers the scene toward the target configuration without requiring a full reset. Here, we show the example video prompt for generating the recovery video of a human index finger poking the purple block for the fourth step of the FMB variant:

\begin{promptbox}[Example Prompt for Non-prehensile Recovery Video Generation]
Use the provided START image as the first frame and the provided GOAL image as the last frame.
The last frame must match the provided goal image: same camera view, same object poses. Single continuous shot, fixed camera, no cuts.

\textbf{Scene:} slightly top-down view of a perforated metal workbench. A red puzzle base is centered. A purple U-shaped block sits on the base. A yellow rectangular bar stays perfectly still. The green and blue pieces behind remain perfectly still and are never touched.

\textbf{Framing:} medium-wide shot with extra margin around the red base; the entire red base is visible with space around it.

\textbf{Contact target (single anchor, no alternatives):} A small solid red star is painted on the outer camera-right-facing vertical side wall of the purple block's right leg, very close to the bottom edge near the red base surface. The center of this red star is the only contact point of the human index finger on the purple block.

\textbf{Hard spatial constraint:
}All visible hand motion stays in the front half of the red base (closest to the bottom of the frame) for the entire shot and never moves into the back half. Make sure the full hand, including the palm and the thumb, is always visible in the view.

\textbf{Hand appearance + stability (reduce DoF):
}It should be a normal adult human hand (realistic proportions), not oversized, not close to the camera. The hand should enter the scene from the right side of the screen.
The index finger extended and the full hand always visible in the view. The fingertip shape and perspective stay consistent (no warping). The finger does not twist or rotate; the pose is stable.

\textbf{Hand approach (straight line, no vertical motion):
}The index finger is low, nearly parallel to the tabletop, and approaches horizontally along a single straight line at the same height as the red star on the purple block. The finger never descends from above, never arcs, and never rises above the purple block's top edge.

\textbf{Continuous-contact rule (no sliding, no switching surfaces):
}The fingertip pad presses onto the center of the red star on the bottom edge of the purple block's right leg and fully covers/occludes the star during contact. From first contact until the purple block is fully seated, the fingertip stays glued to that exact star location (no sliding upward, no rolling over an edge, no moving to the top face, no second contact point). The top face of the purple block is never touched at any time.

\textbf{Action (correct tilt direction, one push only):
}With the fingertip still glued to the red star on the bottom edge of the purple block's right leg, perform one smooth inward push from camera-right to camera-left with a slight downward component. This makes the purple block rotate forward toward the camera (toward the bottom of the frame) like a hinge: the front edge (bottom of frame) goes down into the socket first while the back edge (top of frame) rises slightly. Immediately after this forward tilt, the block drops/slides into the socket underneath and becomes fully seated/flush exactly like the goal image.

\textbf{End:
}As soon as it seats, the finger stops pushing, pauses for a fraction of a second still at the same low height, then retracts straight back to camera-right along the same straight line and exits. No second push, no re-contact elsewhere. Realistic contact physics. Silent.
\end{promptbox}

\section{Geometric Grounding Implementation Details}
\label{app:grounding_details}

\begin{figure*}
    \centering
    \includegraphics[width=\linewidth]{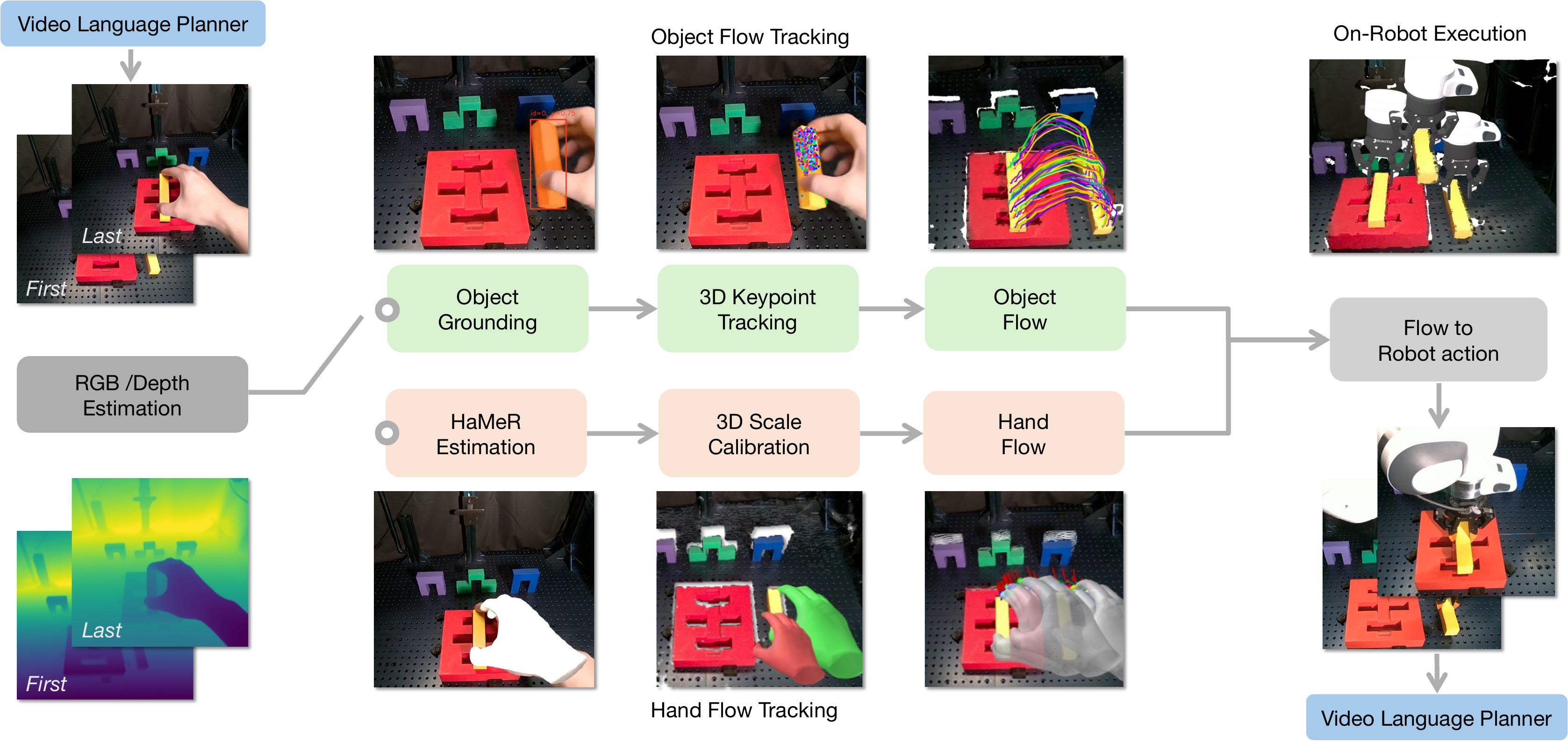}
    \caption{\textbf{\method geometric grounding overview.}
    Given the generated video plan (RGB+Depth), we switch between (top) grounding the target object, tracking 3D keypoints, and recovering object flow, and (bottom) estimating the hand pose with HaMeR, calibrating 3D scale, and computing the hand flow.
    The resulting object/hand flows are converted into robot actions for on-robot execution, with a video language planner enabling closed-loop re-planning and recovery.}
    \label{fig:low-level-system}
\end{figure*}

This section provides implementation details for the geometric grounding module described in \Cref{sec:grounding}. 
Following the components outlined in \Cref{fig:low-level-system}, we describe metric object-flow extraction, hand-flow calibration, contact-constrained grounding for non-prehensile repair, and grasp selection.

\subsection{Metric Depth and Object Flow Extraction}
\label{app:object_flow}

Given a generated RGB video and the target object specified by the planner, we extract a metric object flow by combining monocular depth prediction, temporal depth refinement, object segmentation, and 3D point tracking.

\paragraph{Metric depth recovery.}
We estimate per-frame depth from the generated RGB video using MoGe2~\cite{wang_MoGe2_2025}. 
Given known camera intrinsics, we pass the intrinsics to the model during inference so that the predicted depth is tied to the physical camera geometry.
To reduce frame-to-frame depth flicker, we optionally refine the MoGe2 depth sequence with Consistent Video Depth (CVD)~\cite{luo_Consistent_2020}. 
CVD minimizes optical-flow-induced geometric reprojection errors and regularizes the solution with gradient and normal-consistency terms.
In our implementation, we set the gradient weight $w_{\mathrm{grad}}=2.0$ and the normal-consistency weight $w_{\mathrm{normal}}=5.0$, and we run the optimization at a fixed processing budget of $196608$ pixels.
Because MoGe2 already provides metric depth, we freeze the global shift term during CVD to preserve the metric offset while improving temporal coherence.

\paragraph{RANSAC-based affine calibration.}
To align the generated depth $D_{\mathrm{gen}}$ with the reference sensor depth $D_{\mathrm{sensor}}$ for the first frame, we estimate a global affine transformation:
\begin{equation}
    D_{\mathrm{metric}}(\mathbf{u})
    =
    s_{\mathrm{depth}} D_{\mathrm{gen}}(\mathbf{u})
    +
    t_{\mathrm{depth}},
    \label{eq:depth_affine_calibration_app}
\end{equation}
where $\mathbf{u}$ denotes an image pixel.
We estimate $(s_{\mathrm{depth}}, t_{\mathrm{depth}})$ with RANSAC to reduce sensitivity to invalid sensor measurements, specular regions, and occasional depth prediction failures.

We first build a calibration mask $\mathcal{M}$ that keeps pixels with finite, positive depth in both $D_{\mathrm{gen}}$ and $D_{\mathrm{sensor}}$.
To suppress isolated sensor artifacts, we remove small connected components in the valid-depth mask with area smaller than 50 pixels.
We then run RANSAC for 1000 iterations to find the affine parameters that maximize the number of inliers:
\begin{equation}
\begin{aligned}
(s_{\mathrm{depth}}, t_{\mathrm{depth}})
=
\argmax_{s,t}\;
\sum_{\mathbf{u} \in \mathcal{M}}
\mathbb{I}\!\left(
\left|
sD_{\mathrm{gen}}(\mathbf{u}) + t - D_{\mathrm{sensor}}(\mathbf{u})
\right|
< \tau
\right),
\end{aligned}
\label{eq:depth_ransac_app}
\end{equation}
where $\tau=0.15$ by default.
Finally, we re-fit $(s_{\mathrm{depth}}, t_{\mathrm{depth}})$ using all RANSAC inliers and apply the calibrated parameters to the full generated depth sequence.

\paragraph{3D point tracking.} After depth recovery, we obtain per-frame target-object masks and sample $K$ keypoints on the object in the initial frame.
We track these points through the generated video using a 3D dense point tracker~\cite{zhang_TAPIP3D_2025}, producing the object flow:
\begin{equation}
    \mathcal{F}_{\mathrm{obj}}
    =
    \{
    \mathbf{f}_i^t \in \mathbb{R}^3
    \mid
    i=1,\dots,K,\;
    t=1,\dots,T
    \}.
    \label{eq:object_flow_app}
\end{equation}

\paragraph{Rigid object-motion fitting.}
Given tracked object points at consecutive frames, we estimate the rigid object motion $(\mathbf{R}^t,\mathbf{t}^t)$ by aligning
$\{\mathbf{f}_i^{t-1}\}_{i=1}^{K}$ and $\{\mathbf{f}_i^t\}_{i=1}^{K}$.
Let $\mathbf{c}^{t-1}$ and $\mathbf{c}^{t}$ be the centroids of the two point sets.
We solve for rotation using the Kabsch algorithm~\cite{kabsch_solution_1976}:
\begin{equation}
\mathbf{R}^t
=
\argmin_{\mathbf{R}\in SO(3)}
\sum_{i=1}^{K}
\left\|
\mathbf{R}(\mathbf{f}_i^{t-1}-\mathbf{c}^{t-1})
-
(\mathbf{f}_i^{t}-\mathbf{c}^{t})
\right\|^2.
\label{eq:kabsch_object_flow_app}
\end{equation}
The translation is then computed as
\begin{equation}
    \mathbf{t}^{t}
    =
    \mathbf{c}^{t}
    -
    \mathbf{R}^{t}\mathbf{c}^{t-1}.
    \label{eq:object_translation_app}
\end{equation}
The frame-to-frame transforms are composed to obtain the object trajectory $\{T_{\mathrm{obj}}^t\}_{t=1}^{T}$.

\subsection{Hand Flow Calibration}
\label{app:hand_flow_calibration}

Generated human hands provide useful kinematic priors, but raw hand reconstructions are often not directly metric due to scale hallucination, floating contacts, and projective drift.
We therefore calibrate the reconstructed hand trajectory using object-contact constraints.

\paragraph{Interaction interval.}
We first identify the interval during which the generated hand interacts with the target object.
Assuming the object is static before contact, we define the contact onset $t_{\mathrm{start}}$ as the first frame where the object mask changes relative to the initial mask by more than a threshold $\epsilon$:
\begin{equation}
t_{\mathrm{start}}
=
\argmin_t
\left\{
\frac{
\left|
\mathbf{M}_{\mathrm{obj}}^t
\setminus
\mathbf{M}_{\mathrm{obj}}^{t_0}
\right|
}{
\left|
\mathbf{M}_{\mathrm{obj}}^{t_0}
\right|
}
\ge \epsilon
\right\},
\label{eq:contact_start_mask_app}
\end{equation}
where $\mathbf{M}_{\mathrm{obj}}^t$ is the object mask at frame $t$ and $t_0$ is the first frame.
The contact end $t_{\mathrm{end}}$ is defined analogously as the last frame satisfying the same criterion.
For grasp-mode hand flow, we use $\epsilon=0.9$.

\paragraph{Scale recovery at contact onset.}
At $t_{\mathrm{start}}$, we recover a metric scale that brings the reconstructed hand into contact with the object.
We project the reconstructed hand skeleton into the image plane and select fingertips whose projections overlap with the object mask $\mathbf{M}_{\mathrm{obj}}^{t_{\mathrm{start}}}$.
To allow small projection error, each fingertip projection is represented as a circle with radius 15 pixels.
This produces a candidate contact set $\mathcal{F}_{\mathrm{contact}}$.
For each fingertip $f\in\mathcal{F}_{\mathrm{contact}}$, we estimate an isotropic scale $s_f$ by snapping the fingertip to the center of the corresponding metric object points.
We use
\begin{equation}
    s_{\mathrm{start}}
    =
    \max_{f\in\mathcal{F}_{\mathrm{contact}}} s_f
    \label{eq:hand_scale_start_app}
\end{equation}
as the global hand scale to avoid under-scaling from partially occluded fingertips.
The fingertip attaining this maximum is used as the designated contact finger for subsequent drift compensation.

\paragraph{Drift compensation at release.}
Even after global scale correction, generated videos can exhibit projective drift as the hand moves relative to the camera.
We therefore re-estimate the contact-based scale at release, denoted as $s_{\mathrm{end}}$, and compute a drift offset for the designated contact fingertip:
\begin{equation}
\mathbf{d}_{\mathrm{err}}
=
\mathbf{p}^{t_{\mathrm{end}}}_{\mathrm{tip}}(s_{\mathrm{end}})
-
\mathbf{p}^{t_{\mathrm{end}}}_{\mathrm{tip}}(s_{\mathrm{start}}).
\label{eq:drift_vector_app}
\end{equation}
To avoid perturbing the approach and transport phases, this correction is only applied near release.
We define $t_{\mathrm{corr}}$ as the first frame whose globally scaled fingertip comes within distance $\delta$ of its release configuration:
\begin{equation}
t_{\mathrm{corr}}
=
\argmin_t
\left\{
\left\|
\mathbf{p}^{t}_{\mathrm{tip}}(s_{\mathrm{start}})
-
\mathbf{p}^{t_{\mathrm{end}}}_{\mathrm{tip}}(s_{\mathrm{start}})
\right\|_2
<
\delta
\right\}.
\label{eq:tcorr_app}
\end{equation}
We then linearly ramp the offset over $[t_{\mathrm{corr}}, t_{\mathrm{end}}]$:
\begin{equation}
\begin{aligned}
\alpha(t)
&=
\mathbb{I}(t\ge t_{\mathrm{corr}})
\frac{t-t_{\mathrm{corr}}}{t_{\mathrm{end}}-t_{\mathrm{corr}}},\\
\mathbf{P}^{t}_{\mathrm{cal}}
&=
\mathbf{P}^{t}_{\mathrm{raw}}(s_{\mathrm{start}})
+
\alpha(t)\mathbf{d}_{\mathrm{err}}.
\end{aligned}
\label{eq:hand_drift_ramp_app}
\end{equation}
For grasp-mode hand flow, we use $\delta=5\,\mathrm{cm}$.

\paragraph{SE(3) hand trajectory.}
After calibration, we convert the hand motion into a rigid $SE(3)$ trajectory.
The translation is set to the calibrated 3D position of the designated contact fingertip.
For orientation, we construct a palm frame from 3D hand landmarks: the palm normal is estimated by fitting a plane to the wrist and MCP joints, the in-plane axis is the wrist-to-middle-MCP direction projected onto this plane, and the third axis is their cross product.
This yields a per-frame hand transform $\{T_{\mathrm{hand}}^t\}_{t=1}^{T}$ used for hand-centric grounding.

\paragraph{Rejection criteria.}
Reliable hand trajectory estimation requires the hand to remain largely within the camera view.
We project the estimated hand mesh into the image plane and reject any trajectory where more than 30\% of the hand mask lies outside the image in any frame during the critical interaction phase.
We also reject the trajectory if the primary calibration step fails, e.g., if no fingertip satisfies the projection constraints or the candidate scale set is empty.
In either case, the system discards the current hand flow and prompts the video model for regeneration.

\subsection{Hand-Guided Grasp Selection}
\label{app:grasp_selection}
\method also uses the generated video's human hand as a kinematic prior to select grasps. We implement a three-stage filtering and scoring pipeline to identify the optimal grasp pose.

We first generate $N=5000$ candidate grasps using GraspGen, then filter the candidate grasps using the pipeline below:
\begin{itemize}
    \item \textbf{Contact filtering.}  Given contact fingers $\mathcal{F}_{contact}$ from the hand mesh calibration procedure, we calculate the distance $d_{prox}$ between the grasp’s ``palm line'' (segment connecting gripper finger bases) and these contact points. Grasps with $d_{prox} > 5\,\text{cm}$ are discarded to ensure semantic alignment with the hand-object interaction.

    \item \textbf{Collision avoidance ($S_{collision}$).}  We verify collision-free poses against the scene point cloud (excluding the target object) with a threshold of $1\,\text{mm}$. Grasps that potentially collide with the scene point cloud are filtered.
    
    \item \textbf{Palm line support ($S_{support}$).}  To verify that the gripper makes contact with the object, we compute the fraction of object points that are within $0.5\,\text{cm}$ of the gripper’s opening. This metric is used as a soft filter for the final scoring of the remaining grasps.
\end{itemize}
We rank the unfiltered grasps using a composite score $S_{total}$:
\begin{equation}
    S_{total} = S_{conf} \cdot S_{support}
\end{equation}
where $S_{conf}$ is the raw confidence from GraspGen and $S_{support}$ is the metric representing the fraction of the object surface that could be grasped with the predicted grasp direction.

\subsection{Contact-Constrained Calibration for Non-Prehensile Recovery}
\label{app:nonprehensile_grounding}

For non-prehensile recovery, the recovery prompt specifies both the interaction mode and the contact finger, e.g., ``poke the object with the index finger.''
However, generated recovery videos often distort the contact finger or shift the apparent contact point, making the standard fingertip-overlap calibration unreliable.
We therefore modify the hand-flow calibration routine by explicitly enforcing contact between the prompted fingertip and the selected object surface.

Let $f_{\mathrm{poke}}$ denote the contact finger specified by the recovery prompt.
Unlike grasp-mode calibration, we do not infer the contact finger from fingertip-mask overlap.
Instead, we use the prompted contact finger directly.
At contact onset $t_{\mathrm{start}}$, we estimate a scale and translation correction $(s_{\mathrm{start}}, \Delta \mathbf{t}_{\mathrm{start}})$ such that the prompted fingertip aligns with the metric object surface:
\begin{equation}
    \mathbf{p}_{\mathrm{obj}}^{t_{\mathrm{start}}}
    =
    s_{\mathrm{start}}
    \mathbf{p}_{\mathrm{tip}}^{t_{\mathrm{start}}}
    +
    \Delta \mathbf{t}_{\mathrm{start}},
    \label{eq:nonprehensile_start_contact_app}
\end{equation}
where $\mathbf{p}_{\mathrm{tip}}^{t_{\mathrm{start}}}$ is the reconstructed 3D location of the prompted fingertip and $\mathbf{p}_{\mathrm{obj}}^{t_{\mathrm{start}}}$ is the metric object surface point around the annotated contact region.

We apply this start-anchor correction to the full hand sequence:
\begin{equation}
    \mathbf{P}_{\mathrm{hand,cal}}^t
    =
    s_{\mathrm{start}}
    \mathbf{P}_{\mathrm{hand,raw}}^t
    +
    \Delta \mathbf{t}_{\mathrm{start}}.
    \label{eq:nonprehensile_full_sequence_app}
\end{equation}
As in grasp-mode hand-flow calibration, we compensate residual projective drift near release using a localized ramp correction.
For non-prehensile repair, the drift correction is computed with respect to the prompted contact finger $f_{\mathrm{poke}}$, and we use a smaller threshold $\delta=2\,\mathrm{cm}$ because the object displacement is typically small.
This keeps the fingertip in metric contact with the selected object surface and prevents the recovery trajectory from sliding to a different contact mode.

The remaining steps, including palm-frame orientation estimation and conversion to an end-effector trajectory, are identical to the hand-flow grounding procedure in \Cref{app:hand_flow_calibration}.

\section{Baseline Setups and Oracle Definitions}
\label{app:oracles}
\begin{table}[t]
\centering
\caption{Example oracle subtask sequences provided to NovaFlow, DreamZero, and $\pi_{0.5}$.}
\label{tab:oracle_definitions}
\small
\setlength{\tabcolsep}{6pt}
\renewcommand{\arraystretch}{1.3} 

\newlist{tablist}{enumerate}{1} 
\setlist[tablist]{
    label*=\textbf{\thetablisti}, 
    leftmargin=*,
    nosep,                        itemsep=1.2pt,                  before=\vspace{-0.6\baselineskip}, after=\vspace{0\baselineskip}      }

\begin{tabularx}{\linewidth}{@{} >{\bfseries\raggedright\arraybackslash}p{0.25\linewidth} X @{}}
\toprule
Task & \textbf{Oracle Sub-step Sequence} \\ \midrule

Four-layer Block Stacking & 
    \begin{tablist}
        \item[1.1] Grasp the blue block.
        \item[1.2] Put the blue block onto the red block.
        \item[2.1] Grasp the green block.
        \item[2.2] Put the green block onto the blue block.
        \item[3.1] Grasp the yellow block.
        \item[3.2] Put the yellow block onto the green block.
    \end{tablist} \\
\midrule

Color Sorting & 
    \begin{tablist}
        \item[1.1] Grasp the yellow block.
        \item[1.2] Put the yellow block into the yellow container.
        \item[2.1] Grasp the blue block.
        \item[2.2] Put the blue block into the blue container.
        \item[3.1] Grasp the red block.
        \item[3.2] Put the red block into the red saucer.
    \end{tablist} \\
\midrule

Hidden Object Search & 
    \begin{tablist}
        \item[1.1] Grasp the left drawer's black handle.
        \item[1.2] Pull the black drawer handle straight outward.
        \item[2.1] Grasp the right drawer's green handle.
        \item[2.2] Pull the green drawer handle straight outward.
        \item[3.1] Grasp the block in the right drawer.
        \item[3.2] Place the block onto the top surface of the left drawer.
    \end{tablist} \\

\bottomrule
\end{tabularx}
\vspace{-10pt}
\end{table}

For evaluation, both $\pi_{0.5}$ and DreamZero are evaluated zero-shot from their released checkpoints under the required DROID camera configuration.

We provide $\pi_{0.5}$, NovaFlow, and DreamZero with an ``oracle planner'': each receives a fixed sequence of ground-truth sub-goals in text form, so that their evaluation isolates visual grounding and execution from high-level planning. We present an example in~\Cref{tab:oracle_definitions}.
MOKA receives the task instruction and uses its own VLM-based planner to decompose it into subtasks, without oracle subtask sequences. Its evaluation therefore includes task planning as well as grounding and execution.
VLP-style also produces its own plan and receives no oracle.
It keeps our video generation and our training-free hybrid-flow execution, and replaces only the planner with VLP's beam search over language and video candidates scored by VLM-judged progress toward the final goal.
Because this baseline replans over the remaining horizon whenever execution deviates from the predicted state, rather than repairing a single transition, it costs $2$--$3$ times as many planning calls as \method, with the largest gap on the reactive search task. 

\section{Visualizations of Long-horizon Experiment Rollouts}
\label{app:exp_rollout}
\begin{figure}[H]
    \centering
    \includegraphics[width=0.90\linewidth]{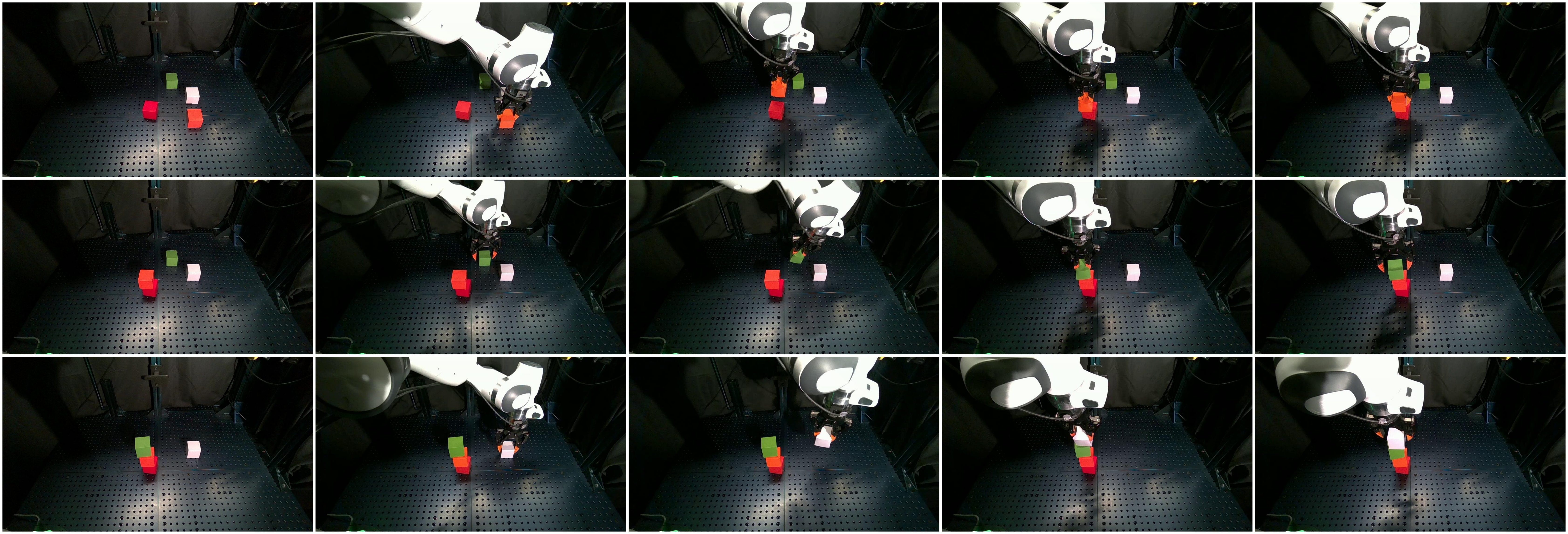}
    \caption{\textbf{Four-layer block stacking experiment.} Each row in this figure illustrates a distinct, sequential step in the block stacking task. The frames in each row display the execution of that specific step, ordered chronologically from left to right. This rollout demonstrates the model's ability to handle precise geometric constraints as the robot identifies, grasps, and aligns multiple blocks vertically.}
    \label{fig:block_stacking_rollout}
    \vspace{-5pt}
\end{figure}

\begin{figure}[H]
    \centering
    \includegraphics[width=0.90\linewidth]{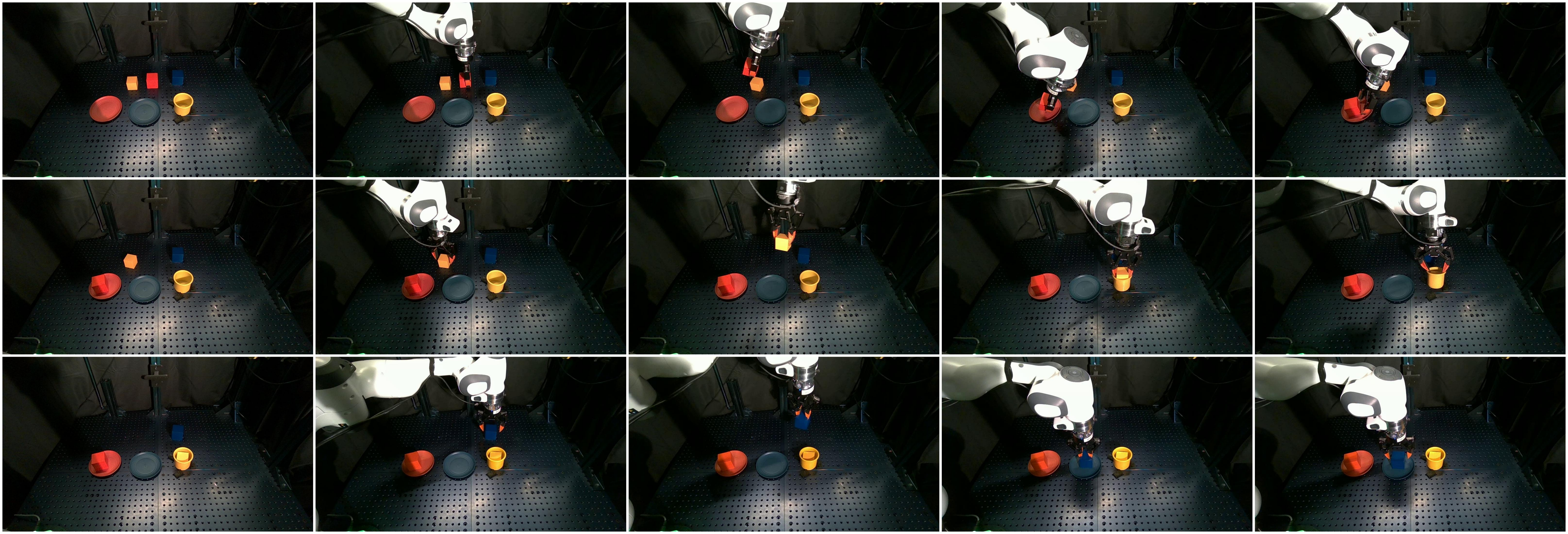}
    \caption{\textbf{Color sorting experiment.} Each row in this figure illustrates a distinct, sequential step in the color sorting task. The frames in each row display the execution of that specific step, ordered chronologically from left to right. This sequence highlights the semantic understanding of the planner as the robot sorts objects into containers with matching colors and the execution precision of preventing the yellow block from getting stuck at the yellow container.}
    \label{fig:color_sorting_rollout}
\end{figure}

\begin{figure}[H]
    \centering
    \includegraphics[width=0.90\linewidth]{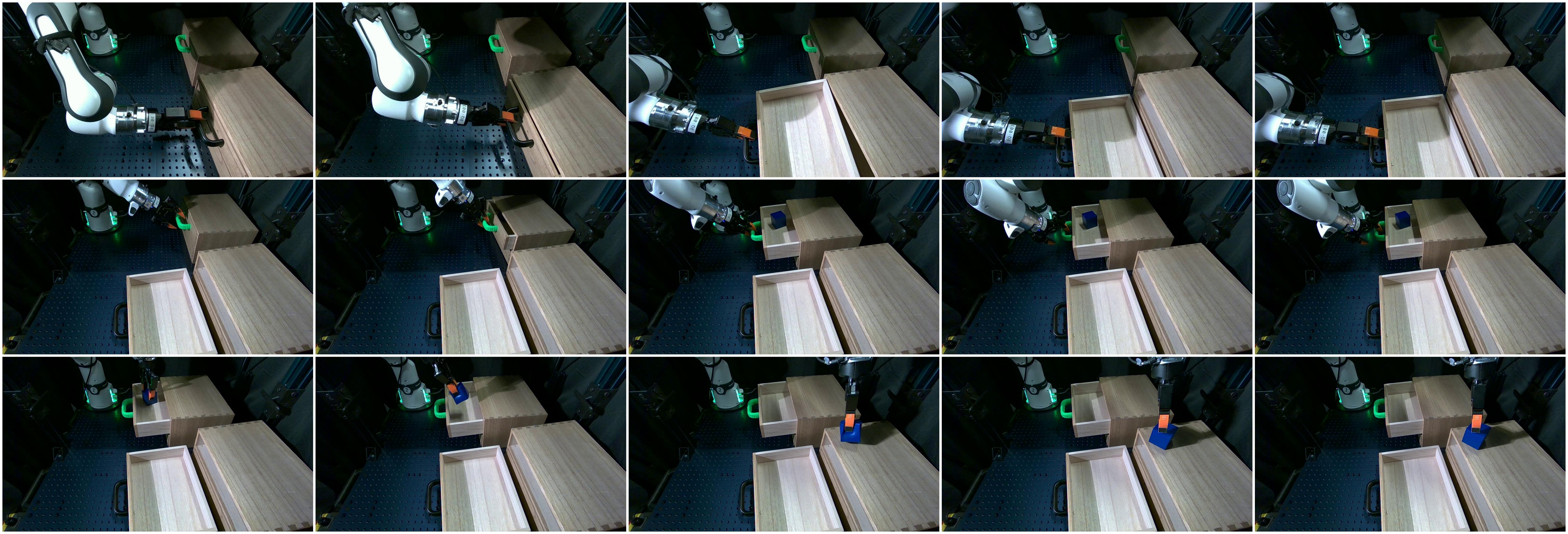}
    \caption{\textbf{Hidden object search experiment.} Each row in this figure illustrates a distinct, sequential step in the hidden object search task. The frames in each row display the execution of that specific step, ordered chronologically from left to right. This rollout shows an example of planning under partial observability where the robot must open drawers to reveal the target object. }
    \label{fig:hidden_object_search_rollout}
\end{figure}

\begin{figure}[H]
    \centering
    \includegraphics[width=0.90\linewidth]{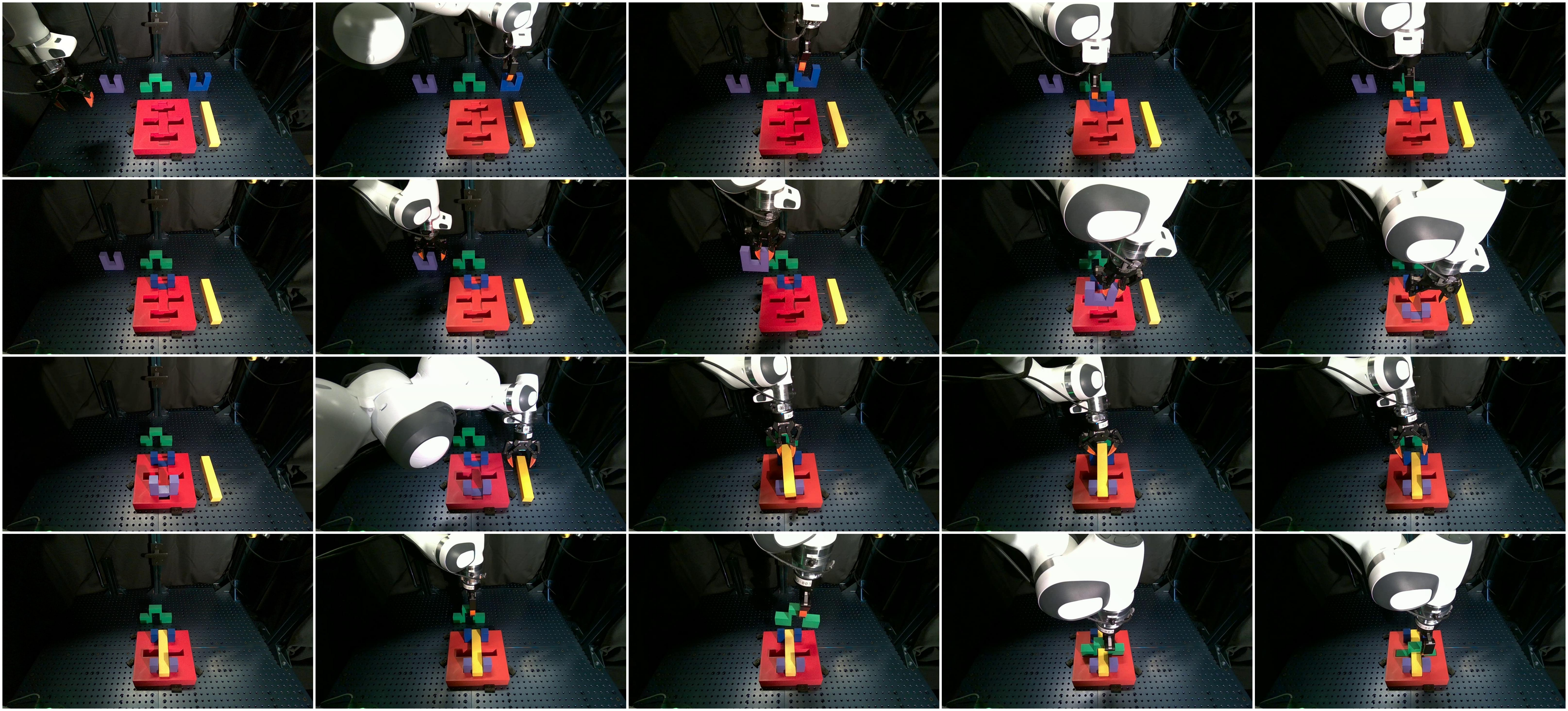}
    \caption{\textbf{Functional Manipulation Benchmark experiment.} Each row in this figure illustrates a distinct, sequential step in an assembly task from the Functional Manipulation Benchmark. The frames in each row display the execution of that specific step, ordered chronologically from left to right. This sequence shows that \method has the capability to solve long-horizon assembly tasks with millimeter precision. }
    \label{fig:fmb_real}
\end{figure}

\begin{figure}[H]
    \centering
    \includegraphics[width=0.90\linewidth]{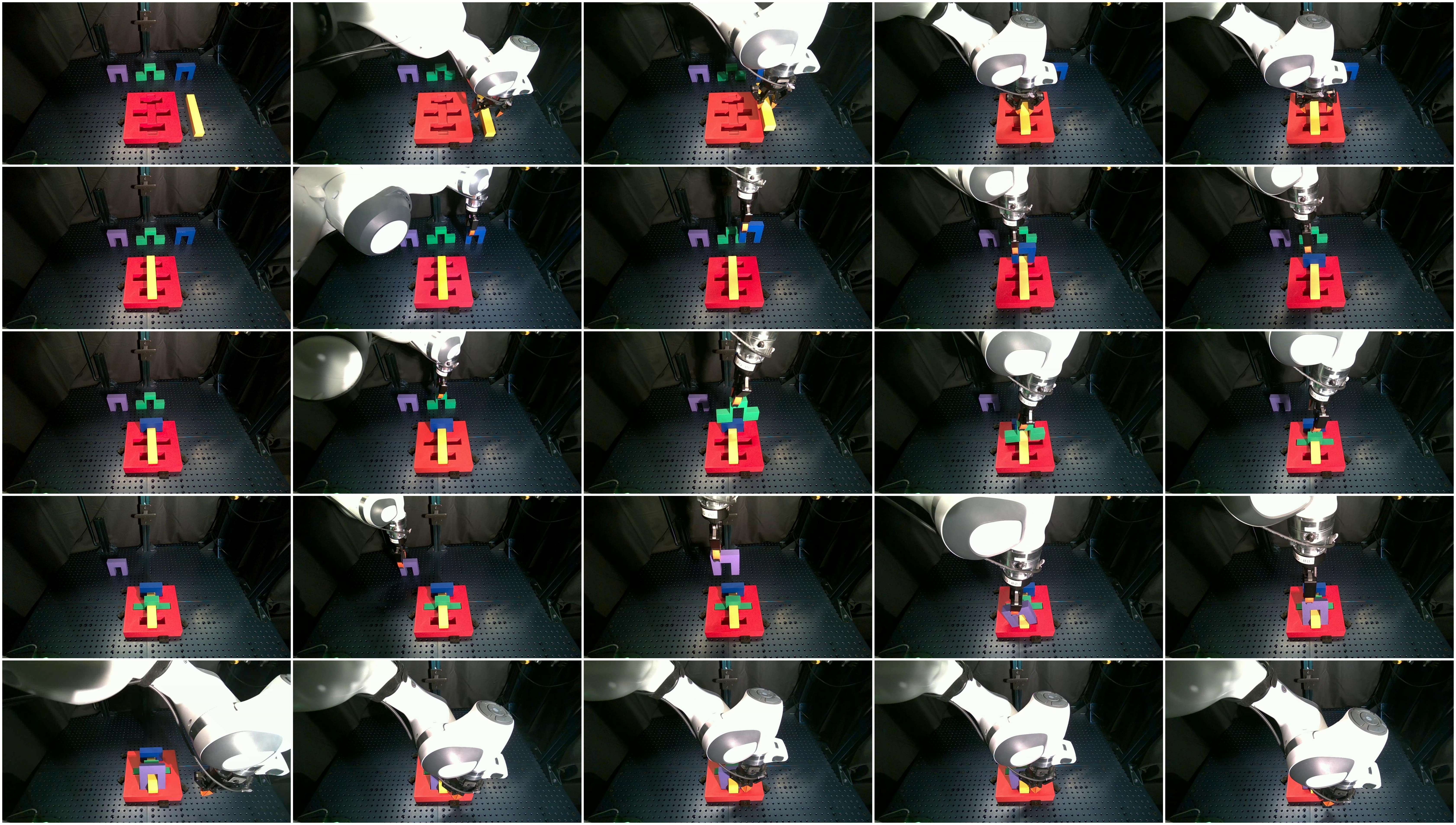}
    \caption{\textbf{Variant of Functional Manipulation Benchmark experiment.} Each row in this figure illustrates a distinct, sequential step in the variant of an assembly task from the Functional Manipulation Benchmark. The frames in each row display the execution of that specific step, ordered chronologically from left to right. The last row of this rollout demonstrates a non-prehensile poking recovery and shows the effectiveness of our hand flow extraction strategy.}
    \label{fig:fmb_fake}
\end{figure}

We provide a qualitative analysis of our video-based planning approach across all long-horizon manipulation tasks. The rollouts from \Cref{fig:block_stacking_rollout} to \Cref{fig:fmb_fake} illustrate that NovaPlan can synthesize coherent multi-step plans and execute them with robustness and precision.

In the Four-layer Block Stacking task (\Cref{fig:block_stacking_rollout}), the planner generates and follows a multi-stage vertical assembly plan, requiring consistent geometric reasoning over several sequential placements.

In the Color Sorting task (\Cref{fig:color_sorting_rollout}), the planner uses semantic cues to match each object to its target container based on visual appearance. It also places the yellow block upright, allowing it to enter the yellow container without jamming at the opening.

In the Hidden Object Search task (\Cref{fig:hidden_object_search_rollout}), the planner operates under partial observability and must first explore the scene (e.g., open multiple drawers if necessary) to reveal the target object before completing the placement.

In the assembly task from the Functional Manipulation Benchmark (\Cref{fig:fmb_real}), \method is able to complete a long-horizon assembly sequence that requires millimeter-level alignment.

Finally, the FMB variant assembly task (\Cref{fig:fmb_fake}) showcases a last non-prehensile poking recovery step, highlighting the effectiveness of our hand-flow extraction strategy for proposing and executing corrective interactions.

% \endgroup

\section{Example Plans and Action Prompts for Video Generation of Long-horizon Tasks}
\label{app:example_plan}
Here, in~\Cref{tab:action_plans}, we present representative multi-step plans produced by the planner. For each long-horizon task, we also provide the corresponding example prompts used to condition the video generation model, together with the object of interest tracked at each step.
\begingroup
\small
\setlength{\tabcolsep}{4pt}
\renewcommand{\arraystretch}{1.18}

\setlength{\LTleft}{\fill}
\setlength{\LTright}{\fill}

\begin{longtable}{@{}
  >{\RaggedRight\arraybackslash}p{\dimexpr0.06\linewidth-1.5\tabcolsep\relax}
  >{\RaggedRight\arraybackslash}p{\dimexpr0.21\linewidth-1.5\tabcolsep\relax}
  >{\RaggedRight\arraybackslash}p{\dimexpr0.59\linewidth-1.5\tabcolsep\relax}
  >{\RaggedRight\arraybackslash}p{\dimexpr0.14\linewidth-1.5\tabcolsep\relax}
@{}}

\caption{Examples of generated multi-step action plans, action prompts for step video generation, and objects to track.}
\label{tab:action_plans} \\
\toprule
\textbf{Step} & \textbf{Proposed Action} & \textbf{Action Prompt for Video Generation} & \textbf{Object to Track} \\
\midrule
\endfirsthead

\multicolumn{4}{c}{\textbf{\tablename\ \thetable{} -- continued from previous page}} \\
\toprule
\textbf{Step} & \textbf{Proposed Action} & \textbf{Action Prompt for Video Generation} & \textbf{Object to Track} \\
\midrule
\endhead

\midrule \multicolumn{4}{r}{\textit{Continued on next page}} \\
\endfoot

\bottomrule
\endlastfoot

\multicolumn{4}{@{}>{\RaggedRight\arraybackslash}p{\linewidth}@{}}{\cellcolor{gray!10}\textbf{Task 1: 4-Layer Block Stacking}} \\*
\multicolumn{4}{@{}>{\RaggedRight\arraybackslash}p{\linewidth}@{}}{\textit{Goal: Stack three blocks onto the other block.}} \\*
1 &
Reach to the orange block, grasp it from the top, lift it, move it to the red block, center it, and lower until seated. &
A human hand reaches to the orange block sitting on the tabletop, grasps it firmly from the top, lifts it straight up, carries it to the red block on the tabletop on its right side, carefully centers it over the red block, and lowers it until fully seated and stable before releasing. The human hand should make every effort not to occlude the orange block during the action and keep the orange block fully visible. The hand completely leaves the frame after the action is done and the overall scene remains still. Do not introduce any new objects in the scene. &
Orange block \\
\addlinespace
2 &
Reach to the green block, grasp it from the top, lift it, move it above the orange block, align it, and lower to seat before releasing. &
A human hand reaches to the green block sitting on the tabletop, grasps it firmly from the top, lifts it quickly, carries it directly above the orange block, puts it onto the orange block's top face, and releases it. The human hand should avoid occluding the green block during the action and keep the green block entirely in view. The hand fully exits the scene after the action is done and the scene stays motionless. Do not create any new objects in the scene. &
Green block \\
\addlinespace
3 &
Reach to the white block, pinch-grasp from the side, lift, hover above the green block, touch the white block's lower edge to align, then release. &
A human hand reaches to the white block sitting on the tabletop, grasps it from the side with a tight pinch, snap-lifts it, carries it quickly to hover just above the green block's top face, touches the white block's lower edge lightly to the green top to align, and releases it. The human hand should take care not to occlude the white block during the action and keep the white block fully in frame. The hand leaves the scene entirely after the action is done and the whole scene stays still. Do not create any new objects in the scene. &
White block \\
\midrule

\multicolumn{4}{@{}>{\RaggedRight\arraybackslash}p{\linewidth}@{}}{\cellcolor{gray!10}\textbf{Task 2: Color Sorting}} \\*
\multicolumn{4}{@{}>{\RaggedRight\arraybackslash}p{\linewidth}@{}}{\textit{Goal: Put objects into containers of matching colors.}} \\*
1 &
Reach to the blue cube, grasp it from the top, lift it, move it above the center of the blue dish, lower it onto the dish, and release inside the boundary. &
A human hand reaches to the blue cube on the tabletop, grasps it from the top, lifts it clear of the surface, moves it above the center of the blue dish, lowers it onto the dish surface, and releases after ensuring it is fully inside the dish boundary. The human hand should make every effort not to occlude the blue cube during the action and keep the blue cube clearly in view. The hand fully exits the scene after the action is done and the scene remains still. &
Blue cube \\
\addlinespace
2 &
Reach to the yellow cube, grasp it from the top, lift it, move it straight above the yellow cup opening, lower it into the cup, and release without contacting the rim. &
A human hand reaches to the yellow cube on the left side of the tabletop in the scene, grasps it from the top, lifts it clear of the surface, moves it straight above the yellow cup opening, lowers it down into the cup until it is fully seated inside, and releases without contacting the rim. The human hand should avoid blocking the yellow cube during the action and keep the yellow cube fully in frame. The hand completely leaves the frame after the action is done and the whole scene stays still. &
Yellow cube \\
\addlinespace
3 &
Reach to the orange cube, grasp it from the top, lift it, move it over the center of the orange dish, lower until it contacts the bottom, and release fully inside. &
A human hand reaches to the orange cube on the tabletop, grasps it firmly from the top, lifts it clear of the surface, moves it straight over the center of the orange dish, lowers it until it contacts the dish bottom, and releases with the cube fully inside the dish boundary. The human hand should take care not to occlude the orange cube during the action and keep the orange cube entirely visible. The hand fully exits the scene after the action is done and the whole scene stays still. &
Orange cube \\
\midrule

\multicolumn{4}{@{}>{\RaggedRight\arraybackslash}p{\linewidth}@{}}{\cellcolor{gray!10}\textbf{Task 3: Hidden Object Search}} \\*
\multicolumn{4}{@{}>{\RaggedRight\arraybackslash}p{\linewidth}@{}}{\textit{Goal: Find the object hidden in one drawer unit and place it on top of the other unit.}} \\*
1 &
Reach to the right unit's black handle, grasp it firmly, and pull the drawer straight outward slowly until the interior is clearly visible. &
A human hand reaches for the right unit’s black handle, grips it firmly, and draws the drawer straight outward slowly until the inside is clearly visible. The hand should make every effort not to occlude the black handle and keep it fully visible throughout. After the drawer is open, the hand exits the frame completely and the scene remains static. Do not introduce any new objects in the scene. &
Black handle \\
\addlinespace
2 &
Reach to the left unit's green handle, grasp it securely, and pull the drawer straight out at a steady pace until the interior is clearly visible. &
A human hand reaches to the left unit's green handle, grasps it firmly, holds it and pulls the drawer outward steadily until the interior is clearly visible. The hand should avoid blocking the green handle and keep it entirely in view during the motion. Once the drawer is opened, the hand fully exits the scene and everything stays still. Do not create any new objects in the scene. &
Green handle \\
\addlinespace
3 &
Place the blue block in the left drawer onto the top surface of the right drawer and release it. &
A human hand grasps the blue block from the top in the left drawer, lifts it up, and moves it onto the top surface of the right drawer and releases it. The human hand should take care not to occlude the blue block during the action and keep the blue block fully in frame. The hand fully exits the scene after the action is done and the whole scene stays still. Do not create any new objects in the scene. &
Blue block \\
\midrule

\multicolumn{4}{@{}>{\RaggedRight\arraybackslash}p{\linewidth}@{}}{\cellcolor{gray!10}\textbf{Task 4: Functional Manipulation Benchmark}} \\*
\multicolumn{4}{@{}>{\RaggedRight\arraybackslash}p{\linewidth}@{}}{\textit{Goal: Insert all colored blocks on the table into their matching sockets in the red base.}} \\*
1 &
Reach to the blue U-shaped block, grasp it near the top edge, lift it, approach the right-side socket from behind the red base, align the inner notch, slide forward while lowering, then press down to seat flush. &
A human hand picks up the blue U-shaped block from the back right side of the tabletop and places it into the furthest horizontal slot at the top of the red board. The open side of the U faces the center. The human hand then fully leaves the scene after the action is done and the whole scene stays still. The human hand should make every effort not to occlude the blue block during the action. &
Blue block \\
\addlinespace
2 &
Reach to the purple U-shaped block, grasp it near the top edge, lift it, align it with the lowest front socket, lower into the slot, and press down to seat flush. &
A human hand picks up the purple U-shaped block from the back left side of the tabletop and inserts it into the lowest horizontal socket at the front of the red board. The open side of the U faces the center, aligning with the blue block. The human hand then fully exits the scene after the action is done and the scene remains still. The human hand should avoid occluding the purple block during the action. &
Purple block \\
\addlinespace
3 &
Reach to the yellow rectangular bar, grasp it at the midpoint, lift it, align it with the long central groove, lower it into the slot, and press until it sits fully seated. &
A human hand picks up the long rectangular yellow block sitting on the right side of the red base on the tabletop and places it into the long central groove on the red base, seating it centered longitudinally within the vertical groove and aligning it with the U-shape openings of the blue and green blocks. The bottom edge of the yellow block extends past the purple block, terminating near the bottom edge of the recessed cross-shaped area of the red base. The human hand leaves the scene entirely after the action is done and the whole scene stays still. The human hand should take care not to occlude the yellow block during the action. &
Yellow bar \\
\addlinespace
4 &
Reach to the green block, grasp it near the top center, lift it, align it with the middle slot, lower and press down until it sits flush and locks over the yellow bar. &
A human hand picks up the green block and inserts it into the middle horizontal slot on the red board. The green block sits in the middle horizontal slot and the arch of the green block straddles directly over the center of the yellow bar, locking it down into the red base. The human hand fully exits the scene after the action is done and the scene remains static. The human hand should make every effort not to occlude the green block during the action. &
Green block \\
\midrule

\multicolumn{4}{@{}>{\RaggedRight\arraybackslash}p{\linewidth}@{}}{\cellcolor{gray!10}\textbf{Task 5: Functional Manipulation Benchmark-Variant}} \\*
\multicolumn{4}{@{}>{\RaggedRight\arraybackslash}p{\linewidth}@{}}{\textit{Goal: Insert all colored blocks on the table into their matching sockets in the red base.}} \\*
1 &
Reach to the yellow rectangular bar, grasp it at the midpoint, lift it, align it with the long central groove, lower it into the slot, and press down until it sits fully flush. &
A human hand holds the long rectangular yellow block and places it into the long central groove running vertically down the center of the red board. Ensure it is lying flat at the bottom of the pocket. The human hand exits the scene completely after the action is done and the whole scene stays still. The human hand should avoid occluding the yellow block during the action. &
Yellow bar \\
\addlinespace
2 &
Reach to the blue block, grasp it near the top edge, lift it, align it with the top horizontal slot, lower it to straddle over the yellow bar, and press down to seat fully. &
A human hand picks up the blue block that is sitting on the right backside of the tabletop and inserts the blue block into the top horizontal slot at the top of the red base. It should straddle over the yellow bar and sit fully in the furthest horizontal slot. The human hand fully leaves the scene after the action is done and the whole scene stays still. The human hand should make every effort not to occlude the blue block during the action. &
Blue block \\
\addlinespace
3 &
Reach to the green block, grasp it near the top center, lift it, align it with the middle slot, lower to bridge over the yellow bar, and press down until fully seated. &
A human hand picks up the green block that is sitting on the tabletop and inserts it into the middle horizontal slot. It should bridge over the yellow bar and sit fully in the middle horizontal slot. The human hand leaves the scene entirely after the action is done and the scene remains still. The human hand should take care not to occlude the green block during the action. &
Green block \\
\addlinespace
4 &
Reach to the purple U-shaped block, grasp it near the top edge, lift it, align it with the lowest remaining socket, lower it into place, and press down to seat flush. &
A human hand picks up the purple U-shaped block that is sitting on the tabletop and inserts it into the last remaining empty horizontal socket at the lowest part of the red board. It should bridge over the yellow bar and sit upright in the lowest horizontal socket in the red base. The human hand fully exits the scene after the action is done and the whole scene stays still. The human hand should avoid occluding the purple block during the action. &
Purple block \\

\end{longtable}
\endgroup

\section{Failure Attribution, Module Ablation, and Prompt Sensitivity}
\label{app:attribution}

This appendix gives the full results of the analyses summarized in \Cref{sec:exp4}.
All numbers use the three long-horizon tasks of \Cref{sec:exp1} with ten trials per task.

\subsection{Module-Level Failure Attribution}
\label{app:failure_attribution}

We record the intermediate outputs of each executed action. A failed action is assigned to the \emph{first} stage whose output cannot be reliably consumed by the next stage.
For example, if depth estimation returns a distorted scene but 3D tracking then produces an unusable trajectory, the failure is attributed to depth estimation rather than to tracking.
Video generation is counted as a failure when the entire sampled set lacks a feasible rollout. The verifier filters individual infeasible rollouts before selection.

\Cref{tab:component_results}(b) reports the resulting distribution.
The highest module failure rates are for 3D object tracking (20.0\%) and monocular depth estimation (12.9\%). Hand flow provides a fallback when the recovered object trajectory is unreliable.
Video generation, object segmentation, hand segmentation, and recovery reasoning have no attributed failures in these trials. Rollout selection and state verification each have a failure rate of 1.2\%. Infeasible candidate videos are filtered and regenerated before execution.

Single-view generation is nonetheless a hard bottleneck for the irregular FMB assembly parts: only $17/64$ and $16/40$ individually generated videos are physically feasible in the variant and standard layouts respectively, which is why we use FMB to probe the boundary of the \method pipeline.

\subsection{Controlled Component Ablation}
\label{app:ablation_protocol}

Each ablation in \Cref{tab:component_results}(a) replaces exactly one module with a simpler alternative while holding every other module, prompt, and geometric threshold fixed.
We evaluate pre-execution outputs by their correctness and execution-time mechanisms by full-task success.

\textbf{Module-level output correctness} applies to modules whose output can be judged directly against the archived trial data, independently of whether the surrounding trial eventually succeeds.
\emph{Rollout verification} measures the fraction of selections that pick a physically feasible rollout; the ablation removes the accumulated object flow image from each candidate tile and scores using frames alone.
\emph{State verification} measures the fraction of correct success/failure judgements; the ablation withholds the generated target frame $I^{\mathrm{target}}_t$, leaving only the pre- and post-execution observations.
\emph{Hand calibration} measures the fraction of hand trajectories that reach metric contact with the object; the ablation uses raw HaMeR output without contact-based scale and drift correction.
\emph{Strategic beam search} measures the fraction of trials yielding a valid action ordering on the coupled tasks; the ablation substitutes purely reactive one-step proposal.

\textbf{Long-horizon task success} measures the effect of execution-time mechanisms on full-task completion.
\emph{Execution re-grounding} is ablated by reusing the rollout selected during planning instead of regenerating from the latest real observation.
\emph{Local recovery} is ablated by terminating the trial on the first verification failure.
\emph{Hybrid grounding} is ablated by disabling the switch to hand flow and executing object flow only.

\subsection{Prompt Sensitivity}
\label{app:prompt_sensitivity}

Because \method conditions its VLM stages on structured prompts, we test whether performance depends on exact phrasing.
We test three prompt variants. The \emph{paraphrase} preserves the original instructions with different wording. The \emph{compressed} prompt shortens these instructions while retaining their structure. The \emph{generic} prompt keeps the input and output schema but removes the physical and causal checks.
Each variant is run three times over the archived examples from the same ten trials per long-horizon task, and we report mean correctness of each VLM stage in \Cref{tab:prompt_sensitivity}.

\begin{table}[H]
\centering
\small
\caption{Prompt sensitivity of each VLM stage (\% correct, mean of three runs).}
\label{tab:prompt_sensitivity}
\setlength{\tabcolsep}{8pt}
\renewcommand{\arraystretch}{1.1}
\begin{tabular}{@{}lcccc@{}}
\toprule
\textbf{VLM role} & \textbf{Original} & \textbf{Paraphrase} & \textbf{Compressed} & \textbf{Generic} \\
\midrule
Action proposal    & \textbf{100.0} & \textbf{100.0} & \textbf{100.0} & 97.6 \\
Rollout selection  & 98.8 & \textbf{100.0} & 97.6 & 78.8 \\
State verification & \textbf{98.8} & 97.6 & 95.3 & 89.4 \\
Recovery reasoning & \textbf{100.0} & \textbf{100.0} & 95.8 & 75.0 \\
\bottomrule
\end{tabular}
\end{table}

Paraphrasing and compression change correctness by at most 4.2$\%$ across the four VLM stages.
Removing the physical and causal checks lowers rollout-selection correctness from 98.8\% to 78.8\% and recovery-reasoning correctness from 100.0\% to 75.0\%. These results indicate that the reasoning instructions contribute more to accuracy than their exact phrasing.

\end{document}